\documentclass[journal]{IEEEtran}
\ifCLASSINFOpdf
\else
\fi
\usepackage{graphicx}
\usepackage[numbers,sort&compress]{natbib}

\usepackage{textcomp}
\usepackage{times}
\usepackage{epsfig}
\usepackage{amsmath}
\usepackage{amssymb}
\usepackage{multirow}
\usepackage{booktabs}
\usepackage{makecell}
\usepackage{subcaption}
\usepackage[table]{xcolor}
\newcommand{\etal}{\textit{et al}.}

\definecolor{redlinkcolor}{rgb}{0.79607843, 0.25098039, 0.25882353}
\definecolor{bluecitecolor}{rgb}{0,0.36,0.69}
\usepackage[colorlinks=true,linkcolor=redlinkcolor,citecolor=bluecitecolor,urlcolor=bluecitecolor]{hyperref}

\begin{document}
%
\title{Towards Better De-raining Generalization via \\ Rainy Characteristics Memorization and Replay}
%
%
%
\author{Kunyu Wang,
        Xueyang Fu,
        Chengzhi Cao,
        Chengjie Ge,
        Wei Zhai,
        and~Zheng-Jun Zha,~\IEEEmembership{Member,~IEEE}
\thanks{The authors are with the School of Information Science and Technology, University of Science and Technology of China, Hefei 230026, China (e-mail: kunyuwang@mail.ustc.edu.cn; xyfu@ustc.edu.cn; chengzhicao@mail.ustc.edu.cn; chengjiege@mail.ustc.edu.cn; wzhai056@ustc.edu.cn; zhazj@ustc.edu.cn).}}

\markboth{IEEE TRANSACTIONS ON NEURAL NETWORKS AND LEARNING SYSTEMS}%
{IEEE TRANSACTIONS ON NEURAL NETWORKS AND LEARNING SYSTEMS}
 
%



\maketitle

\begin{abstract}
Current image de-raining methods primarily learn from a limited dataset, leading to inadequate performance in varied real-world rainy conditions. To tackle this, we introduce a new framework that enables networks to progressively expand their de-raining knowledge base by tapping into a growing pool of datasets, significantly boosting their adaptability. Drawing inspiration from the human brain's ability to continuously absorb and generalize from ongoing experiences, our approach borrow the mechanism of the complementary learning system. Specifically, we first deploy Generative Adversarial Networks (GANs) to capture and retain the unique features of new data, mirroring the hippocampus's role in learning and memory. Then, the de-raining network is trained with both existing and GAN-synthesized data, mimicking the process of hippocampal replay and interleaved learning. Furthermore, we employ knowledge distillation with the replayed data to replicate the synergy between the neocortex's activity patterns triggered by hippocampal replays and the pre-existing neocortical knowledge. This comprehensive framework empowers the de-raining network to amass knowledge from various datasets, continually enhancing its performance on previously unseen rainy scenes. Our testing on three benchmark de-raining networks confirms the framework's effectiveness. It not only facilitates continuous knowledge accumulation across six datasets but also surpasses state-of-the-art methods in generalizing to new real-world scenarios. 
\end{abstract}


%
\IEEEpeerreviewmaketitle

\section{Introduction}
\begin{figure}[t]
    \centering
    \includegraphics[width=\linewidth]{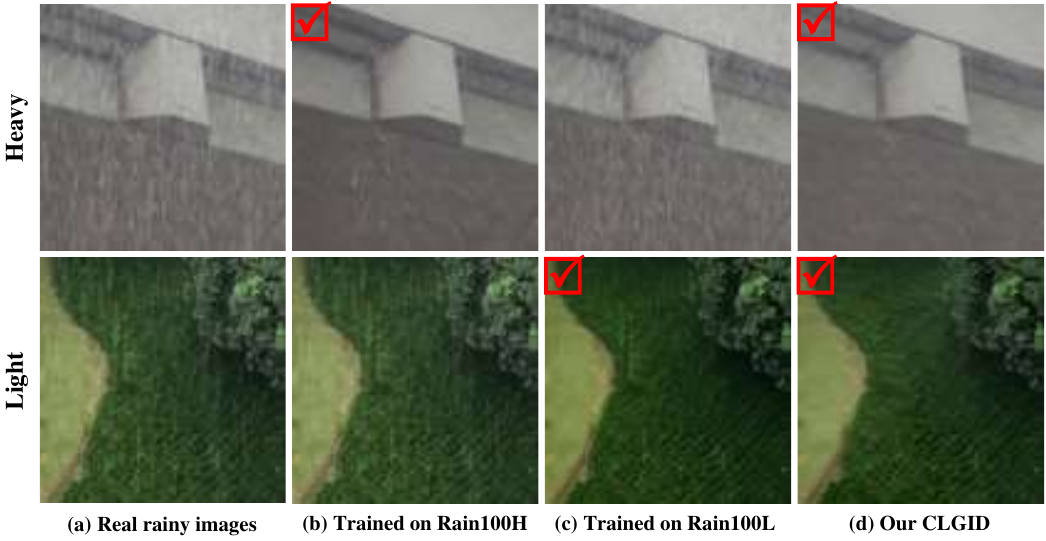}
    \caption{Visual comparison between MFDNet \cite{wang2023multi}, trained on a fixed de-raining dataset, and employing our CLGID to accumulate de-raining knowledge from increasingly abundant datasets. The red check mark indicates the network's promising performance. While MFDNet trained on a fixed dataset can only handle specific types of real-world rain streaks (i.e., rain100H$\rightarrow$Heavy, rain100L$\rightarrow$Light), our CLGID is capable of addressing various real-world rainy scenes.}
    \vspace{-0.3cm}
    \label{example} 
\end{figure}

\IEEEPARstart{S}{ingle} image de-raining, which seeks to eliminate rain streaks from images to reveal their clean versions, is pivotal for enhancing the efficacy of subsequent vision tasks like classification and detection \cite{wang2023generalized,wang2024towards,wang2023centernet,xu2023rssformer, xu2022instance,xu2021dc,zhang2024uni,zhang2024navid}. Despite the advancements achieved by deep learning-based de-raining methods \cite{fu2017clearing,zamir2021multi,ren2019progressive,jiang2020multi,wang2020rethinking,li2019heavy,jiang2021multi,huang2022contrastive}, a critical shortcoming remains: their reliance on learning specific de-raining patterns from a limited set of rainy images. This approach results in underwhelming performance when applied in varied real-world conditions, due to the inability to fully represent the complexity of real-world rain distribution, as illustrated in Fig~\ref{example}. To surmount this challenge, it is essential to develop de-raining methods capable of continually expanding their knowledge by learning from an ever-increasing collection of de-raining datasets. This strategy enables the networks to significantly improve their adaptability and performance in real-world scenarios, addressing the partial coverage issue of current specific de-raining mappings.

One potential solution is to integrate newly acquired data with existing data and retrain the network from scratch. However, this approach requires retraining for each new dataset, and as the combined dataset grows, the retraining costs escalate, making this method impractical due to the significant computational expense. Alternatively, the de-raining network could be sequentially trained on newly acquired data. However, this approach is prone to catastrophic forgetting \cite{parisi2019continual}, which occurs due to interference between new and previously learned knowledge, resulting in an ineffective accumulation of de-raining knowledge.

\begin{figure}[t]
	\centering
	\subfloat[\label{fig:a}]{
		\includegraphics[width=0.8\linewidth]{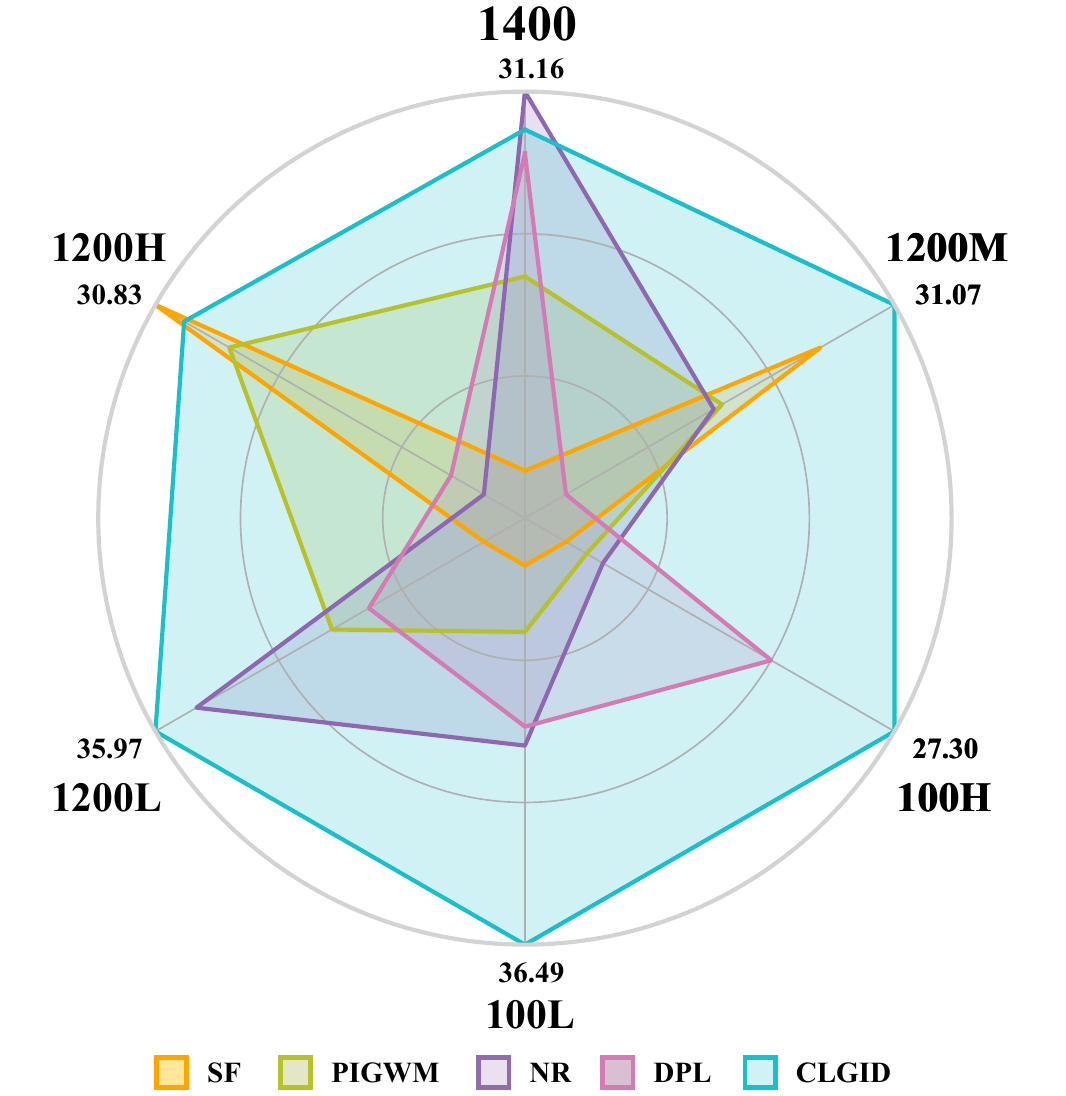}}
        \\
        \centering
	\subfloat[\label{fig:a}]{
		\includegraphics[width=0.9\linewidth]{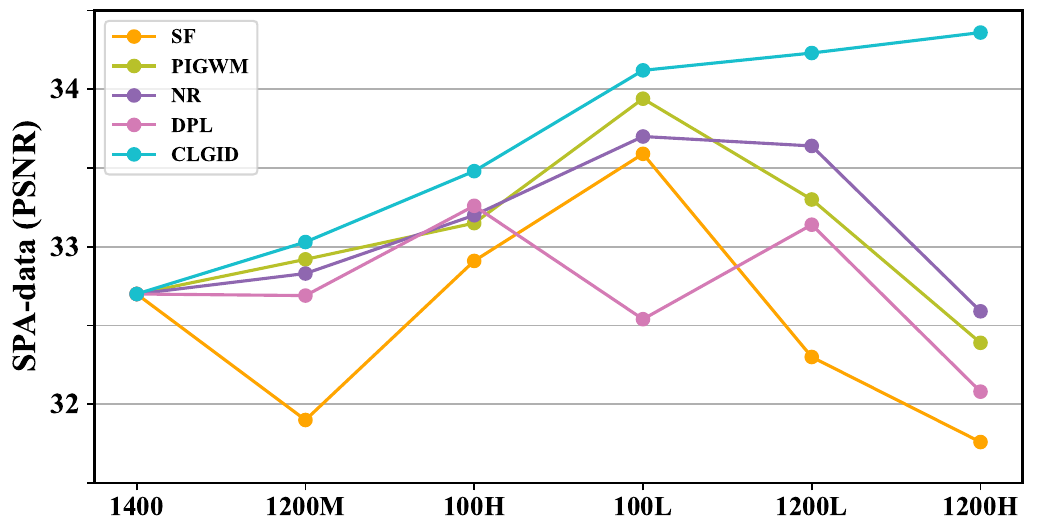}}
	\caption{{(a) Memory performance on individual datasets after training on the six-dataset stream 1400-1200M-100H-100L-1200L-1200H. (b) Generalization performance on unseen SPA-data during training.} }
        \vspace{-0.3cm}
	\label{fig2} 
\end{figure}

To address this issue, recent efforts have been focused on efficiently gathering de-raining insights from data streams. These initiatives include various strategies, such as adjusting model parameter weights~\cite{zhou2021image}, segregating model parameters~\cite{xiao2021improving}, and learning prompts specific to the task~\cite{liu2023dual}. However, these methods continue to face challenges in memory capacity and generalization. As illustrated in Fig~\ref{fig2}, the introduction of additional datasets leads to a saturation of memory capacity, which in turn limits the enhancement of the network's ability to generalize.

On the other hand, humans have an exceptional capacity to continuously learn and remember various events, extracting statistical patterns from these events to develop the ability to generalize to new situations. The complementary learning system of the human brain, involving the hippocampus and neocortex, is pivotal in this cognitive process~\cite{parisi2019continual,mcclelland1995there,kumaran2016learning}. Inspired by this remarkable capability, an intriguing question arises: can we draw inspiration from the human brain's complementary learning system, which facilitates ongoing memory of events and generalization across these memories, to overcome existing challenges of the image de-raining?

In this paper, we draw inspiration from the human brain's Complementary Learning system to introduce a new Generalized Image De-raining framework (CLGID). Specifically, the complementary learning system comprises of the hippocampus and the neocortex. The hippocampus is responsible for learning and storing unique perceptions of events \cite{mcnaughton1987hippocampal}. It then replays these memories, mixed with new events, to the neocortex. Through a cycle of replaying hippocampal memories and interspersed learning, the neocortex harmonizes these memories with existing  neocortical knowledge \cite{french1999catastrophic}, gradually extracting structured insights and developing the ability to generalize to new scenarios.

Mirroring this, we use GANs to mimic the hippocampus by learning and store the rainy characteristics of each dataset. The de-raining network, acting as the neocortex, is then trained on a mix of GAN-generated memories and current data. This method replicates the hippocampal to neocortical replay and interleaved learning process, fostering the network's ability to generalize across data. Additionally, we incorporate knowledge distillation with replayed data to ensure the neocortical activity patterns, triggered by hippocampal replays, align with pre-existing neocortical knowledge. 
Extensive experiments on three representative de-raining networks~\cite{wang2023multi,zamir2022restormer,zamir2021multi} confirm that CLGID effectively preserves memory across six datasets and continuously improves generalization to unseen real-world images, outperforming existing methods. {Fig.~\ref{fig2} presents results based on MFDNet~\cite{wang2023multi}.}

Additionally, recent discoveries in cognitive science \cite{mcclelland2013incorporating,tse2011schema} bolster the theory of the complementary learning system, suggesting the neocortex can swiftly assimilate structured knowledge when new events closely resemble past ones. Motivated by these insights, we propose a pattern similarity-based training acceleration algorithm to enhance our CLGID framework. Essentially, we evaluate the similarity in rain patterns between the new dataset and GAN-generated memories before beginning training. Depending on this similarity, we modify the number of training cycles for the new dataset. A higher similarity leads to fewer training cycles and reduced training duration. This strategy allows us to cut down the overall training time by an average of $48\%$ without sacrificing generalization capabilities.

To summarize, the key contributions of our paper are outlined below:
\begin{itemize}
\item Inspired by the complementary learning system in human brain, we propose a novel continual learning paradigm for image de-raining, emphasizing human-like knowledge accumulation, offering a novel perspective on enhancing model generalization.
\item We propose a generalized de-raining framework (CLGID) that integrates generative replay, interleaved training, and consistency-based distillation to accumulate de-raining knowledge across multiple datasets while alleviating catastrophic forgetting.
\item We introduce a similarity-based training speedup mechanism that reduces training iterations for new datasets based on their similarity to previously learned ones, expediting training without compromising generalization.
\end{itemize}

The structure of the following sections is organized as such: Section  \ref{section:Related} provides a comprehensive review of relevant literature. Section \ref{section:Methodology} delves into the intricacies of the proposed CLGID framework. Subsequently, Section \ref{section:Experiments} examines various ablation studies and presents the experimental findings related to memory and generalization capabilities. Finally, Section \ref{section:Conclusion} provides concluding remarks on this work.

\section{Related works}
\label{section:Related}
\subsection{Single Image De-raining}
Recent years have witnessed significant progress in image de-raining. Most traditional methods for addressing this problem employ kernels \cite{kim2013single}, low-rank approximation \cite{chen2013generalized,chang2017transformed}, and dictionary learning \cite{kang2011automatic,huang2013self,luo2015removing,wang2017hierarchical}. However, due to using the hand-crafted features to estimate the rain model in traditional de-raining methods, they fail under complex rain conditions and produce degraded image contents. Recently, deep learning-based methods \cite{yasarla2020syn2real,wei2019semi,liu2021unpaired,ye2022unsupervised,jin2019unsupervised,zheng2020single,chang2023direction,wang2020rain} have emerged for rain streak removal and achieved impressive restoration performance.
Wei \etal ~\cite{wei2019semi} proposed a semi-supervised transfer learning method, which exploits statistics prior to aligning the synthetic and real-world domains.
AirNet~\cite{li2022all} proposed a contrastive-based all-in-one restoration framework that handles diverse unknown corruptions without requiring prior degradation information, showing strong flexibility in real-world scenarios. 
Fu \etal ~\cite{fu2019lightweight} introduced a lightweight deep network with fewer than 8K parameters, yet it achieved competitive performance in image deraining tasks, making it well-suited for applications on mobile devices.
CLEARER~\cite{gou2020clearer} introduced a NAS-based multi-scale architecture that adaptively balances performance and complexity, replacing handcrafted design. 
Rajeev \etal ~\cite{yasarla2020syn2real} proposed a Gaussian-based semi-supervised learning framework, which involves iteratively training on the labeled synthetic data and unlabeled real-world data for better generalization.
MaIR~\cite{li2024mair} leveraged a Mamba-based structure with continuity-preserving scanning and sequence attention, achieving state-of-the-art results across multiple restoration tasks, including de-raining.
Yang \etal ~\cite{yang2019scale} proposed a recurrent wavelet learning approach that can effectively remove rain streaks from images, even in cases of heavy rain accumulation under low light conditions.
DPCNet~\cite{he2024dual} introduced a dual-path spatial-frequency interaction network with adaptive fusion, effectively restoring rain-corrupted details across diverse conditions. 
Ren \etal ~\cite{ren2020single} proposed a single recurrent network and a bilateral recurrent network for rain streak removal in images.
Yang \etal ~\cite{yang2020removing} introduced a fractal band learning network for rain streak removal, employing low-order constructed modules as the fundamental units of high-order ones, capturing potential hierarchical dependencies among band features.
DLINet~\cite{li2023efficient} adopted a decoupled architecture for rain location and intensity, mitigating feature interference and improving subtask specialization. 
Zhang \etal ~\cite{zhang2021dual} proposed a dual attention-in-attention model to simultaneously remove both rain streaks and raindrops.
DualCNN~\cite{sivaanpu2022dual} introduced a dual-branch network for joint raindrop and rain streak removal, combining detail restoration and color enhancement with guided filtering and skip connections.
Wang \etal ~\cite{wang2021rain} proposed a comprehensive bayesian generative model for rainy images, wherein the rain layer is parameterized as a generator with inputs representing physical rain factors. This model contributes to enhancing the performance of de-raining networks and mitigates the necessity for extensive pre-collection of training samples.
Zhu \etal ~\cite{zhu2020learning} developed a gated non-local deep residual learning framework for image de-raining, avoiding over de-raining or under de-raining caused by global residual learning in existing de-raining networks.
PLSA~\cite{liu2024pixel} presented a 3DLUT-based enhancement framework with pixel-adaptive intensity modeling and saturation-aware correction, improving visibility and perceptual quality in degraded or low-light regions.
Hu \etal ~\cite{hu2021single} presented a depth-guided non-local module embedded in a deep neural network for real-time single-image rain removal. The proposed module captures long-term dependencies between feature positions in a depth-guided manner, allowing for effective removal of rain streaks and fog.
Wang \etal ~\cite{wang2023rcdnet} introduced the rain convolutional dictionary network, which embeds intrinsic priors of rain streaks, thereby enhancing both interpretability and performance in de-raining tasks.
More deep learning-based methods can be found in \cite{li2021comprehensive,yang2020single}.

\subsection{Accumulating De-raining Knowledge}
While existing de-raining methods have yielded promising results, the majority of them are primarily focused on training their networks using a fixed set of synthetic or real data. However, relying on de-raining networks to exclusively learn specific rain mappings from a fixed dataset is inadequate for comprehensively addressing the intricate and diverse rain distribution encountered in real-world rainy images. Confronted with unseen real-world rainy scenarios, the de-raining network often exhibits a significant decline in performance. Therefore, it is imperative to empower de-raining networks to unremittingly accumulate de-raining knowledge from increasingly abundant datasets, rather than depending solely on a static dataset, facilitating de-raining networks in constantly acquiring generalization ability in real-world scenarios.
Recently, researchers have been dedicated to endowing de-raining networks with the aforementioned capability.
Zhou \etal ~\cite{zhou2021image} developed the parameter importance guided weights modification method to enable the de-raining networks to learn from a sequence of synthetic datasets. 
Xiao \etal ~\cite{xiao2021improving} explored an neural reorganization method to ensure the accumulation of de-raining knowledge and overcome catastrophic forgetting.
Gu \etal ~\cite{gu2023incremental} proposed a memory management strategy called associative memory that associates the current pathway with the historical representation, enabling incremental rain removal.
Liu \etal ~\cite{liu2023dual} proposed a novel prompt learning-based continual learning scheme for single image de-raining, which handles different types of rain streaks with a single model.
However, these methods can effectively accumulate de-raining knowledge only across a limited number of datasets. With the influx of more datasets, the memory capacity tends to become saturated, impeding the accumulation of de-raining knowledge and hindering the constant improvement of network generalization.
In contrast, our proposed CLGID method not only maintains efficient memory capacity but also achieves constant improvement of generalization ability with the arrival of more datasets, resulting in superior generalization performance on unseen real-world data.

\begin{figure*}[htbp]
\begin{center}
\includegraphics[width=0.95\linewidth]{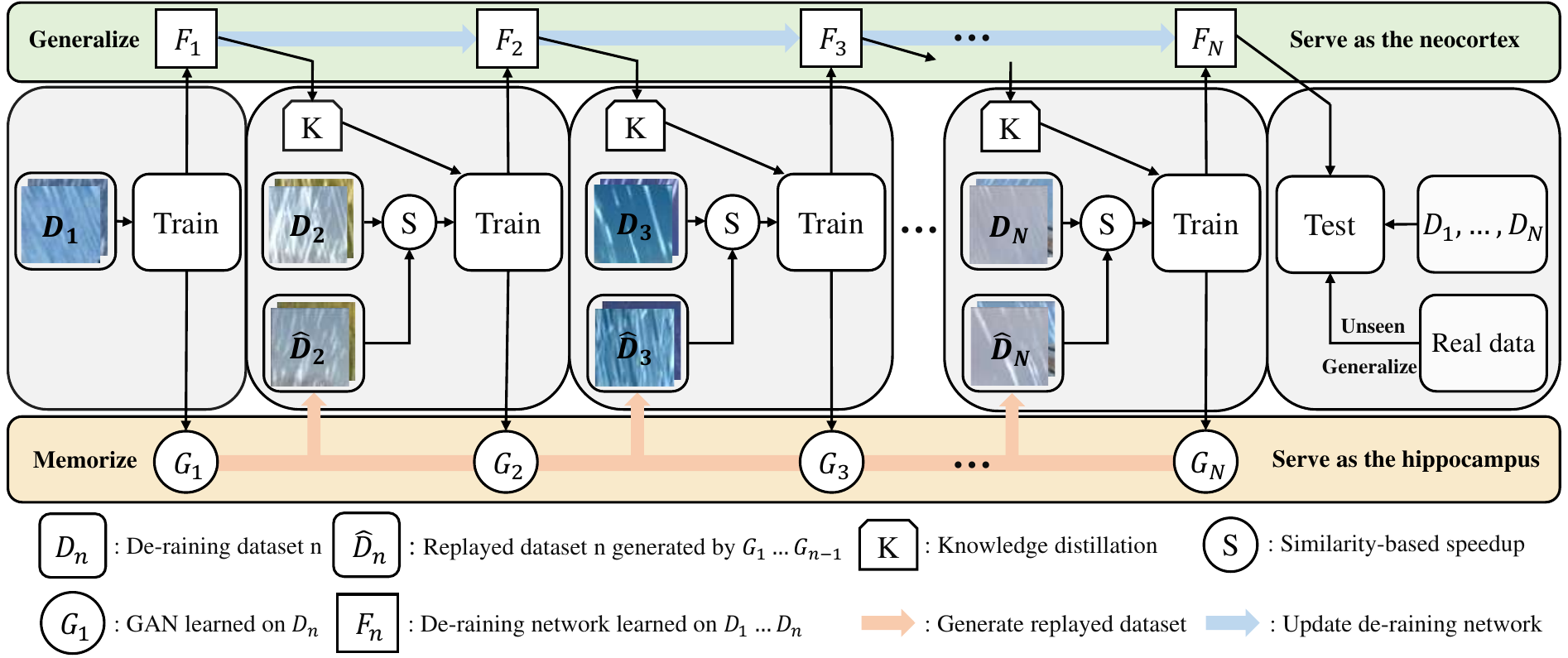}
\end{center}
\caption{{Flowchart of the proposed CLGID framework. Given each incoming dataset $D_n$ , a corresponding GAN ($G_n$) mimics the hippocampus to learn and store rain characteristics. Previously trained GANs replay past data ($\hat{D}_n$), interleaved with current data ($D_n$), to train the de-raining network ($F_n$), analogous to the neocortex. Knowledge distillation ensures consistency between current and past knowledge. A similarity-based training speedup algorithm further reduces training iterations.}}
\vspace{-0.5cm}
\label{Method}
\end{figure*}

\section{Methodology}
\label{section:Methodology}
Given a stream of de-raining datasets $\{ D_i \}_{i=1}^N$, where N is the total number of the datasets and each dataset $D_i$ contains $M_i$ pairs of rainy images $x_i$ and clean background $y_i$, our goal is to constantly improve the de-raining network generalization by accumulating de-raining knowledge from increasingly abundant datasets. To achieve this goal, we seek inspiration from the human brain. Humans possess the ability to constantly learn and memorize a stream of perceived events, extracting statistical structures across memorized events to acquire the generalization ability to unseen situations. The complementary learning system, comprised of the hippocampus and neocortex, significantly contributes to the aforementioned process. Inspired by the complementary learning system in the human brain, we propose a new generalized image de-raining framework, CLGID, to improve the de-raining network generalization by borrowing the mechanism of the complementary learning system. The detailed flowchart of CLGID is shown in Fig.~\ref{Method}.

\subsection{Imitating the complementary learning system}
In the complementary learning system, the hippocampus allows for the learning and individualized storage of a stream of perceived event.
Accordingly, we adopt GANs, as exemplified by VRGNet \cite{wang2021rain}, to learn and store the rain streak characteristics for each incoming dataset individually, which play the role of the hippocampus. Specifically, with the arrival of a new dataset $D_{n\in [1,N]}$, a corresponding GAN $G_n$ will be trained on $D_{n}$ to learn the rain streak characteristics of $D_{n}$. The forward propagation process of learning the rain streak characteristics can be formulated as:
\begin{equation}
    \alpha_n, \beta_n = R_{i} \left(x_n ; W_{R_{i}}\right),
\end{equation}
\begin{equation}
    z_n \leftarrow \operatorname{Reparameterize}(\alpha_n, \beta_n),
\end{equation}
\begin{equation}
    r_n = R_{g} \left(z_n ; W_{R_{g}}\right),
\end{equation}
where $R_{i}$ and $R_{g}$ indicate the rain inference network and rain generator in VRGNet, $W_{R_{i}}$ and $W_{R_{g}}$ denote the parameters of $R_{i}$ and $R_{g}$, $\alpha_n$ and $\beta_n$ are the posterior parameters (i.e., mean and variance, respectively) of latent variable $z_n$ inferenced by $R_{i}$, $r_n$ represents the rain streak layer of $x_n$ generated by $R_{g}$. More training details can be found in \cite{wang2021rain}. Consequently, we have obtained the $G_n$, which store the rain streak characteristics of $D_n$.

 Then, the hippocampus will repeatedly replay the memorized events in the hippocampus, interleaved with the new events, back to the neocortex. 
 To imitate the hippocampus-to-neocortex replay and the interleaved learning, we first construct a replayed dataset $\hat{D}_n$ generated by previously learned GANs $\{G_1, G_2, \cdots, G_{n-1}\}$ that are trained on $\{D_1, D_2, \cdots, D_{n-1}\}$, respectively. 
 Then, the replayed dataset $\hat{D}_n$ and the new dataset $D_n$ are utilized for training the de-raining network $F_n$, which serves as the neocortex. 
 Specifically, for replayed dataset $\hat{D}_n$, we determine which GAN generates each pair of images by uniformly sampling from the previously learned GANs, which can be expressed as:
 \begin{equation}
     G_r \leftarrow \{ G_1, G_2, \cdots, G_{n-1} \}.
 \end{equation}
 Moreover, we sample the latent variable $\hat{z}_n$ from isotropic Gaussian distribution and use the rain generator $R_{g}$ of $G_r$ to generate the rain streak layer $\hat{r}_n$ \cite{wang2021rain}:
 \begin{equation}
	\hat{z}_n \leftarrow \mathcal{N}(0,\textbf{I}_t),
 \label{5}
\end{equation}
 \begin{equation}
	\hat{r}_n = R_{g}(\hat{z}_n; W_{R_{g}}),
 \label{6}
\end{equation}
where $\textbf{I}_t \in \mathbb{R}^{t \times t}$ is the unit matrix. For each generated rain layer $\hat{r}_n$, we randomly select a clean background image $y_n$ from $D_n=\{ x_n^{m}, y_n^{m} \}_{{m}=1}^{M_n}$ to form a replayed rainy image $\hat{x}_n$ by adding $\hat{r}_n$ to $y_n$. Therefore, the replayed dataset $\hat{D}_n$ can be recorded as:
\begin{equation}
\hat{D}_{n}=\left\{y_{n}^{m}+\hat{r}_{n}^{m}, y_{n}^{m}\right\}_{m=1}^{M_n} \triangleq \left\{\hat{x}_{n}^{m}, \hat{y}_{n}^{m}\right\}_{m=1}^{M_n}.
\label{7}
\end{equation}

In addition, throughout this iterative process of replaying hippocampal memories and interleaved learning, the neocortical activity patterns activated by the hippocampus’ replayed events remain consistent with existing neocortical knowledge. To imitate this characteristic, we employ knowledge distillation with replayed data to ensure the consistency of the knowledge in the de-raining network. Specifically, we distill the de-raining knowledge from the previous obtained de-raining network $F_{n-1}$ that trained with the arrival of $D_{n-1}$ to the current de-raining network $F_{n}$. We send the replayed rainy images of $\hat{D}_{n}$ to both $F_{n-1}$ and $F_{n}$ and ensure the consistency of the knowledge by encouraging the outputs of $F_{n-1}$ and $F_{n}$ to be similar.
Thus, the neocortex gradually extracts structured knowledge across events and acquires the ability to generalize to unseen situations. 

To sum up, the total loss function of the proposed framework comprises the interleave loss, which includes the new loss and the replay loss, along with the consistency loss, which enables the de-raining network to constantly acquire generalization ability to unseen real-world data after training on a stream of de-raining datasets:
\begin{equation}
    \mathcal{L}_{\text{new}} = \mathcal{L}_{\text{char}}(F_{n}(x_n), y_n) + \mathcal{L}_{\text{edge}}(F_{n}(x_n), y_n),
\end{equation}
\begin{equation}
    \mathcal{L}_{\text{replay}} = \mathcal{L}_{\text{char}}(F_{n}(\hat{x}_n), \hat{y}_n) + \mathcal{L}_{\text{edge}}(F_{n}(\hat{x}_n), \hat{y}_n),
\end{equation}
\begin{equation}
    \mathcal{L}_{\text{interleave}} = \mathcal{L}_{\text{replay}} + \mathcal{L}_{\text{new}},
\end{equation}
\begin{equation}
    \mathcal{L}_{\text{consist}} = \left\|F_{n}\left(\hat{x}_n\right)-{F}_{n-1}\left(\hat{x}_n\right)\right\|_1,
\end{equation}
\begin{equation}
    \mathcal{L}_{\text{total}} = \mathcal{L}_{\text{interleave}}+\lambda \times \mathcal{L}_{\text{consist}},
\end{equation}
where the $\mathcal{L}_{\text{char}}$ and $\mathcal{L}_{\text{edge}}$ are the loss functions of the de-raining network, as exemplified by MPRNet \cite{zamir2021multi}, $\lambda$ is the hyper-parameter for balancing the interleave loss and the consistency loss, the new loss and the replay loss are considered to be of equal significance as the size of the new dataset ${D}_{n}$ is the same as the replayed dataset $\hat{D}_{n}$.
{The complexity analysis of the proposed CLGID framework is provided in Appendix \ref{complexity}, detailing how its parameter count, FLOPs, and time cost evolve as the number of datasets increases.}

\subsection{Similarity-based training speedup algorithm}
Recent neuroscience studies \cite{mcclelland2013incorporating,tse2011schema} have updated the complementary learning system theory, demonstrating that the neocortex can extract structured knowledge across events faster than originally suggested if new events are highly similar to previously learned events. Inspired by these findings, we design a similarity-based training speedup algorithm to complement our CLGID framework.
We calculate the rain characteristics similarity between the new and previously learned datasets and reduce the total training iterations and training time for the new dataset according to its similarity to the previously learned datasets. The greater the similarity, the fewer the total training iterations and the shorter the training time. Specifically, with the arrival of new dataset $D_n=\{ x_n^{m}, y_n^{m} \}_{{m}=1}^{M_n}$, we utilize the replayed dataset \(\hat{D}_n\) constructed for training \(D_n\), which is composed by uniformly sampling from all previously trained GANs \(\{ G_{1}, G_{2}, \cdots, G_{n-1} \}\). To compute similarity, we divide \(\hat{D}_n\) into subsets \(\{ \hat{D}_{1}, \hat{D}_{2}, \cdots, \hat{D}_{n-1} \}\) according to GAN sources and extract Histogram of Oriented Gradients (HOG)~\cite{dalal2005histograms} from the rainy images in both \(\hat{D}_{i\in \{1, \dots, n-1\}}\) and \(D_n\) as:
\begin{equation}
\label{13}
    \hat{h}_{i} = \text{HOG}(\hat{x}_{i}), \quad
    {h}_{n} = \text{HOG}({x}_{n}).
\end{equation}
The Kullback-Leibler (KL) divergence~\cite{kullback1951information} is then used to compute the similarity between \(\hat{D}_i\) and \(D_n\):
\begin{equation}
    s_i^n = D_{KL}( \hat{h}_{i} || {h}_{n}),
\end{equation}
where \(s_i^n\) denotes the similarity coefficient between \(\hat{D}_{i}\) and \(D_n\). After calculating all similarity coefficients $\{ s_1^n, s_1^n, \cdots, s_{n-1}^n \}$, we select the smallest of these values as the final similarity coefficient $S_n$ between the new dataset $D_{n}$ and the previously learned datasets:
\begin{equation}
    S_n = \min_{i \in \{1, \dots, n-1\}}{s_i^n}.
\end{equation}
Since the value range of KL divergence is $[0, +\infty]$, we utilize a mapping function to map its value range from $[0, +\infty]$ to $[0, 1]$ for bounding its value domain:
\begin{equation}
    \widehat{S}_n = 1-e^{-S_n}.
    \label{17}
\end{equation}
Finally, we reduce the total training iterations from $I_n$ to $\widehat{I}_n$ for the new dataset ${D}_{n}$ based on the final similarity coefficient $\widehat{S}_n$, thus reducing the training time:
\begin{equation}
    \widehat{I}_n = \widehat{S}_n \times I_n.
\end{equation}

\subsection{{Improving scalability of the framework}}
\noindent{\textbf{Similarity-Based Selective GAN Training:}
Current framework trains one GAN for each new dataset, which leads to linearly growing training and storage cost. Inspired by the similarity-based training speedup mechanism, we extend this idea to GAN training and determine whether a new GAN should be trained for the incoming dataset $D_n=\{ x_n^{m}, y_n^{m} \}_{{m}=1}^{M_n}$. 
Following the similarity computation pipeline defined in Eq. \ref{13}–\ref{17}, we first compute the similarity $s_{i}^{n}$ between $D_n$ and each GAN-generated replayed dataset $\hat{D}_{i \in \{1,\dots,n-1\}}$.
Let $S_n = \min_{i \in \{1, \dots, n-1\}} s_{i}^{n}$, we denote the normalized divergence between the current dataset $D_n$ and all replayed datasets $\{\hat{D}_i\}_{i=1}^{n-1}$ as $\hat{S}_n = 1 - e^{-s_n} \in [0, 1]$. A higher value of $\hat{S}_n$ implies greater dissimilarity. }

{To improve the scalability of GAN training, we train a new GAN $G_n$ only when the normalized similarity score $\hat{S}_n$ exceeds a predefined threshold $\hat{T} \in (0, 1)$:
\begin{equation}
    \text{Train } G_n \quad \text{if } \hat{S}_n > \hat{T}.
\end{equation}
Let the binary indicator be defined as:
\begin{equation}
    \delta_n = \begin{cases}
    1, & \text{if } \hat{S}_n > \hat{T}, \\
    0, & \text{otherwise},
    \end{cases}
\end{equation}
then the total number of trained GANs after $N$ stages is:
\begin{equation}
    |\mathcal{G}_N| = \sum_{i=1}^{N} \delta_i.
\end{equation}
With the increasing number of trained GANs, the diversity of captured rain patterns becomes more comprehensive, lowering the probability of requiring additional GAN training. As a result, both GAN training and storage costs grow sub-linearly, thus improving the overall scalability of the framework.}

\noindent{\textbf{GAN-replayed Data Reuse:}
In current framework, each new dataset $D_n$ requires generating a replayed dataset $\hat{D}_i$ of the same size by uniformly sampling from all previously trained GANs $\{G_i\}_{i=1}^{n-1}$. This leads to $M_n$ GAN forward passes at every stage, and a total replay cost of $\mathcal{O}(MN)$, which scales linearly with the number of datasets and poses a practical limitation.}

{To address this scalability bottleneck, we propose a replay data reuse mechanism. The core idea is to reuse replayed samples generated in previous stages and only perform incremental generation if necessary. Specifically, at stage $n$, the replayed dataset $\hat{D}_n$ is constructed by generating $M_n / (n-1)$ samples using each of the previously trained GANs $\{G_1, \dots, G_{n-1}\}$, where $M_n$ is the size of the current dataset $D_n$. We denote $r_{i,n} = M_n / (n - 1)$ as the required number of replayed samples from $G_i$ at stage $n$. Meanwhile, let $c_{i,n-1} = M_{n-1} / (n - 2)$ be the number of cached samples generated by $G_i$ during the previous stage $n - 1$. Then, for each $i = \{1, \dots, n - 2\}$, we reuse the cached samples and generate only the difference:
\[
\Delta_{i,n} = \max(0, r_{i,n} - c_{i,n-1}).
\]
For the newly introduced GAN $G_{n-1}$, which was not involved in $\hat{D}_{n-1}$, we generate all its required samples:
\[
\Delta_{n-1,n} = r_{n-1,n} = \frac{M_n}{n-1}.
\]
To analyze the total replay cost up to stage $N$, we consider the total number of additional forward passes across all stages:
\begin{align}
C_N &= \sum_{n=2}^{N} \left( \sum_{i=1}^{n-2} \Delta_{i,n} + \Delta_{n-1,n} \right) \notag \\
    &= \sum_{n=2}^{N} (n-2) \cdot \max\left(0, \frac{M_n}{n-1} - \frac{M_{n-1}}{n-2}\right) + \frac{M_n}{n-1}.
\end{align}
As shown by the mathematical proof in Appendix \ref{math}, we can conclude that:
\begin{equation}
    C_N = \mathcal{O}(M \log N),
\end{equation}
demonstrating that the total replay cost grows logarithmically with the number of datasets, thereby enhancing scalability over the original linear design.}

\section{Experiments}
\label{section:Experiments}
In this section, we first introduce the experimental settings. Then, we compare our method with several SOTA methods on a stream of four datasets. Furthermore, we extend the length of the data stream to encompass six datasets to further evaluate the effectiveness of our method in terms of memory and generalization to unseen real rainy images. 
Finally, we conduct comprehensive ablation studies covering the similarity-based training speedup algorithm, the scalability of GANs in the framework, the effectiveness of individual framework components, the impact of hyper-parameter choices, the stability and the convergence behavior of the framework.

\subsection{Experimental Settings}
\textbf{Datasets.}
We adopt six de-raining datasets as the source data for training, including Rain100H \cite{yang2017deep}, Rain100L \cite{yang2017deep}, Rain1400 \cite{fu2017removing}, Rain1200L \cite{zhang2018density}, Rain1200M \cite{zhang2018density}, and Rain1200H \cite{zhang2018density}.
Rain100H and Rain100L, each contain 1,800 training images and 200 testing images. Rain1400 includes 12,600 training images and 1,400 testing images, covering fourteen types of streak orientations and magnitudes. As for Rain1200L, Rain1200M, and Rain1200H, they represent the light-density, medium-density, and heavy-density subsets of Rain1200 based on rain-density labels, with each subset containing 4,000 training and 400 testing images.
For assessing memory performance, after training on a stream of datasets, each method will be tested on the test set of every dataset within the stream to evaluate its memory performance for each dataset.
For assessing generalization performance, following \cite{xiao2021improving}, we choose SPA-data \cite{wang2019spatial} and Real-Internet \cite{wang2019spatial} for qualitative and quantitative evaluation. These two datasets consist of real-world data that has never been seen during training. Furthermore, they exhibit both cross-dataset gap and syn2real gap characteristics, posing challenges to network generalization.
SPA-data contains 29,500 high-quality rain/clean image pairs, divided into 28,500 for training and 1,000 for testing, generated from 170 real rain videos. We utilize the test set for evaluation. Real-Internet comprises 146 real-world images collected from the Internet without ground truth annotations.

\begin{table*}[htbp]
  \centering
  \caption{Qualitative comparison of memory and generalization performance after training on a stream of four datasets in four distinct sequences. The comparisons include the baseline SF, state-of-the-art methods PIGWM \cite{zhou2021image}, NR \cite{xiao2021improving}, and DPL \cite{liu2023dual}, as well as our proposed CLGID, all using MFDNet \cite{wang2023multi} as the de-raining network. $\text{CLGID}^{\dag}$ represents the accelerated training version of CLGID, using our proposed similarity-based training speedup algorithm. \textcolor{black!20}{Individual} signifies training and testing on each dataset individually, providing a reference for evaluating the memory performance. We evaluate generalization on SPA-data, \textit{\textbf{which has never been seen during training}}. We highlight the best results using \underline{\textbf{such}} formatting.}
  \setlength{\heavyrulewidth}{1.3pt}
  \setlength{\tabcolsep}{0.15cm}
  \renewcommand{\arraystretch}{0.95}
  \resizebox{0.9\linewidth}{!}{
    \begin{tabular}{cccccccccc|cc|cc}
    \toprule[\heavyrulewidth]
    \multirow{2}[2]{*}{Training Sequence} & \multirow{2}[2]{*}{Methods} & \multicolumn{2}{c}{Rain1400} & \multicolumn{2}{c}{Rain1200M} & \multicolumn{2}{c}{Rain100H} & \multicolumn{2}{c}{Rain100L} & \multicolumn{2}{c}{Avg Memory} & \multicolumn{2}{c}{SPA-data} \\
    \cmidrule{3-14}          &       & PSNR  & SSIM  & PSNR  & SSIM  & PSNR  & SSIM  & PSNR  & SSIM  & PSNR  & SSIM  & PSNR  & SSIM \\
    \midrule
    \multirow{7}[1]{*}{1400-1200M-100H-100L} & \textcolor{black!20}{Individual} & \textcolor{black!20}{31.79}  & \textcolor{black!20}{0.920}  & \textcolor{black!20}{32.24}  & \textcolor{black!20}{0.920}  & \textcolor{black!20}{27.70}  & \textcolor{black!20}{0.886}  & \textcolor{black!20}{36.16}  & \textcolor{black!20}{0.978}  & \textcolor{black!20}{31.97}  & \textcolor{black!20}{0.926} & \textendash & \textendash \\
          & SF    & 28.67  & 0.878  & 27.00  & 0.833  & 25.54  & 0.840  & 36.34  & 0.977  & 29.39  & 0.882 & 33.59 & 0.941 \\
          & PIGWM & 29.18  & 0.893  & 27.74  & 0.963  & 26.38  & 0.863  & 33.85  & 0.965  & 29.29  & 0.921 & 33.94 & 0.947 \\
          & NR    & 30.21  & 0.908  & 27.95  & 0.874  & 18.39  & 0.615  & 31.74  & 0.939  & 27.07  & 0.834 & 33.70 & 0.947 \\
          & DPL   & 29.64  & 0.876  & 28.21  & 0.875  & 25.57  & 0.844  & 34.61  & 0.962  & 29.51  & 0.889 & 32.54 & 0.931 \\
\cmidrule{2-14}          
          & CLGID & 31.26  & 0.915  & 31.65  & 0.905  & 28.07  & 0.891  & 36.77  & 0.979  & \underline{\textbf{31.94}}  & \underline{\textbf{0.923}} & \underline{\textbf{34.12}} & \underline{\textbf{0.948}}  \\
          & $\text{ CLGID}^{\dag}$ & 31.09  & 0.912  & 30.65  & 0.893  & 27.94  & 0.889  & 36.65  & 0.979  & 31.58  & 0.918 & 34.06  & 0.947 \\
    \midrule[\heavyrulewidth]
    \multirow{2}[2]{*}{Training Sequence} & \multirow{2}[2]{*}{Methods} & \multicolumn{2}{c}{Rain1400} & \multicolumn{2}{c}{Rain1200M} & \multicolumn{2}{c}{Rain100H} & \multicolumn{2}{c}{Rain100L} & \multicolumn{2}{c}{Avg Memory} & \multicolumn{2}{c}{SPA-data} \\
    \cmidrule{3-14}          &       & PSNR  & SSIM  & PSNR  & SSIM  & PSNR  & SSIM  & PSNR  & SSIM  & PSNR  & SSIM  & PSNR  & SSIM \\
    \midrule
    \multirow{7}[1]{*}{1400-100L-1200M-100H} & \textcolor{black!20}{Individual} & \textcolor{black!20}{31.79}  & \textcolor{black!20}{0.920}  & \textcolor{black!20}{36.16}  & \textcolor{black!20}{0.978}  & \textcolor{black!20}{32.24}  & \textcolor{black!20}{0.920}  & \textcolor{black!20}{27.70}  & \textcolor{black!20}{0.886}  & \textcolor{black!20}{31.97}  & \textcolor{black!20}{0.926} & \textendash & \textendash \\
          & SF    & 28.61  & 0.879  & 36.19  & 0.976  & 26.85  & 0.837  & 28.57  & 0.900  & 30.06  & 0.898 & 33.70 & 0.943  \\
          & PIGWM & 29.59  & 0.899  & 33.35  & 0.964  & 29.22  & 0.884  & 26.51  & 0.865  & 29.67  & 0.903 & 33.48 & 0.948 \\
          & NR    & 30.23  & 0.904  & 31.22  & 0.938  & 28.64  & 0.871  & 19.96  & 0.691  & 27.51  & 0.851 & 32.20 & 0.940 \\
          & DPL   & 29.34  & 0.886  & 33.27  & 0.961  & 28.07  & 0.869  & 24.47  & 0.839  & 28.79  & 0.889 & 33.89 & 0.948 \\
\cmidrule{2-14}          & CLGID & 31.24  & 0.916  & 36.21  & 0.978  & 31.74  & 0.908  & 27.68  & 0.887  & \underline{\textbf{31.72}}  & \underline{\textbf{0.922}} & \underline{\textbf{34.42}}  & \underline{\textbf{0.952}}   \\
          & $\text{ CLGID}^{\dag}$ & 31.06  & 0.906  & 36.27  & 0.978  & 30.05  & 0.883  & 27.66  & 0.884  & 31.26  & 0.913 & 34.24  & 0.949  \\
    \midrule[\heavyrulewidth]
        \multirow{2}[2]{*}{Training Sequence} & \multirow{2}[2]{*}{Methods} & \multicolumn{2}{c}{Rain1400} & \multicolumn{2}{c}{Rain1200M} & \multicolumn{2}{c}{Rain100H} & \multicolumn{2}{c}{Rain100L} & \multicolumn{2}{c}{Avg Memory} & \multicolumn{2}{c}{SPA-data} \\
    \cmidrule{3-14}          &       & PSNR  & SSIM  & PSNR  & SSIM  & PSNR  & SSIM  & PSNR  & SSIM  & PSNR  & SSIM  & PSNR  & SSIM \\
    \midrule
    \multirow{7}[1]{*}{100L-100H-1400-1200M} & \textcolor{black!20}{Individual} & \textcolor{black!20}{36.16}  & \textcolor{black!20}{0.978}  & \textcolor{black!20}{27.70}  & \textcolor{black!20}{0.886}  & \textcolor{black!20}{31.79}  & \textcolor{black!20}{0.920}  & \textcolor{black!20}{32.24}  & \textcolor{black!20}{0.920}  & \textcolor{black!20}{31.97}  & \textcolor{black!20}{0.926} & \textendash & \textendash \\
          & SF    & 24.62  & 0.819  & 13.02  & 0.369  & 28.98  & 0.891  & 32.40  & 0.922  & 24.76  & 0.750 & 29.09 & 0.917  \\
          & PIGWM & 26.08  & 0.872  & 14.02  & 0.435  & 28.40  & 0.891  & 30.85  & 0.895  & 24.84  & 0.773 & 28.16 & 0.908 \\
          & NR    & 31.98  & 0.951  & 17.13  & 0.564  & 29.81  & 0.898  & 28.01  & 0.869  & 26.73  & 0.821 & 32.55 & 0.933 \\
          & DPL   & 31.71  & 0.953  & 20.70  & 0.752  & 29.95  & 0.902  & 29.35  & 0.875  & 27.93  & 0.871 & 32.99 & 0.940 \\
\cmidrule{2-14}          & CLGID & 35.62  & 0.975  & 26.09  & 0.847  & 31.36  & 0.920  & 32.00  & 0.914  & \underline{\textbf{31.27}}  & \underline{\textbf{0.914}} & \underline{\textbf{34.20}}  & \underline{\textbf{0.951}}  \\
          & $\text{ CLGID}^{\dag}$ & 34.82  & 0.972  & 25.69  & 0.823  & 31.18  & 0.920  & 31.93  & 0.916  & 30.91  & 0.908 & 34.09  & 0.949  \\
    \midrule[\heavyrulewidth]
        \multirow{2}[2]{*}{Training Sequence} & \multirow{2}[2]{*}{Methods} & \multicolumn{2}{c}{Rain1400} & \multicolumn{2}{c}{Rain1200M} & \multicolumn{2}{c}{Rain100H} & \multicolumn{2}{c}{Rain100L} & \multicolumn{2}{c}{Avg Memory} & \multicolumn{2}{c}{SPA-data} \\
    \cmidrule{3-14}          &       & PSNR  & SSIM  & PSNR  & SSIM  & PSNR  & SSIM  & PSNR  & SSIM  & PSNR  & SSIM  & PSNR  & SSIM \\
    \midrule
    \multirow{7}[1]{*}{100H-100L-1400-1200M} & \textcolor{black!20}{Individual} & \textcolor{black!20}{27.70}  & \textcolor{black!20}{0.886}  & \textcolor{black!20}{36.16}  & \textcolor{black!20}{0.978}  & \textcolor{black!20}{31.79}  & \textcolor{black!20}{0.920}  & \textcolor{black!20}{32.24}  & \textcolor{black!20}{0.920}  & \textcolor{black!20}{31.97}  & \textcolor{black!20}{0.926} & \textendash & \textendash \\
          & SF    & 12.98  & 0.365  & 24.93  & 0.826  & 28.83  & 0.887  & 32.37  & 0.922  & 24.78  & 0.750 & 29.12 & 0.917 \\
          & PIGWM & 14.58  & 0.461  & 26.14  & 0.869  & 29.08  & 0.901  & 30.76  & 0.894  & 25.14  & 0.781 & 28.16 & 0.908 \\
          & NR    & 23.66  & 0.802  & 31.73  & 0.945  & 29.76  & 0.898  & 29.16  & 0.879  & 28.58  & 0.881 & 32.55 & 0.933 \\
          & DPL   & 20.84  & 0.736  & 32.17  & 0.941  & 28.32  & 0.872  & 30.69  & 0.883  & 28.01  & 0.858 & 32.99 & 0.940 \\
\cmidrule{2-14}          & CLGID & 26.74  & 0.851  & 35.54  & 0.974  & 31.11  & 0.919  & 31.90  & 0.914  & \underline{\textbf{31.32}}  & \underline{\textbf{0.915}} & \underline{\textbf{33.96}} & \underline{\textbf{0.949}} \\
          & $\text{ CLGID}^{\dag}$ & 26.29  & 0.832  & 35.18  & 0.971  & 31.10  & 0.918  & 31.78  & 0.911  & 31.09  & 0.908 & 33.82 & 0.948 \\
    \bottomrule
    \end{tabular}%
    }
    \vspace{-0.4cm}
  \label{tab:41}%
\end{table*}%

\begin{table*}[ht]
  \centering
  \caption{Qualitative comparison of memory performance after training on a stream of four datasets in four distinct sequences. All methods utilize Restormer \cite{zamir2022restormer} as the de-raining network. $\text{CLGID}^{\dag}$ represents the accelerated training version of CLGID, using our proposed similarity-based training speedup algorithm. \textcolor{black!20}{Individual} signifies training and testing on each dataset individually, providing a reference for evaluating the memory performance.  We evaluate generalization on SPA-data, \textit{\textbf{which has never been seen during training}}. We highlight the best results using \underline{\textbf{such}} formatting.}
  \setlength{\heavyrulewidth}{1.3pt}
  \setlength{\tabcolsep}{0.15cm}
  \renewcommand{\arraystretch}{0.95}
  \resizebox{0.9\linewidth}{!}{
    \begin{tabular}{cccccccccc|cc|cc}
    \toprule[\heavyrulewidth]
   \multirow{2}[2]{*}{Training Sequence} & \multirow{2}[2]{*}{Methods} & \multicolumn{2}{c}{Rain1400} & \multicolumn{2}{c}{Rain1200M} & \multicolumn{2}{c}{Rain100H} & \multicolumn{2}{c}{Rain100L} & \multicolumn{2}{c}{Avg Memory} & \multicolumn{2}{c}{SPA-data} \\
    \cmidrule{3-14}          &       & PSNR  & SSIM  & PSNR  & SSIM  & PSNR  & SSIM  & PSNR  & SSIM  & PSNR  & SSIM  & PSNR  & SSIM \\
    \midrule
    \multirow{7}[1]{*}{1400-1200M-100H-100L} & \textcolor{black!20}{Individual} & \textcolor{black!20}{32.01}  & \textcolor{black!20}{0.929}  & \textcolor{black!20}{32.80}  & \textcolor{black!20}{0.925}  & \textcolor{black!20}{29.87}  & \textcolor{black!20}{0.914}  & \textcolor{black!20}{38.33}  & \textcolor{black!20}{0.985}  & \textcolor{black!20}{33.25}  & \textcolor{black!20}{0.938} & \textendash & \textendash \\
          & SF    & 26.63  & 0.842  & 21.79  & 0.707  & 21.94  & 0.679  & 38.52  & 0.986  & 27.22  & 0.804 & 32.87 & 0.931  \\
          & PIGWM & 27.80  & 0.875  & 23.61  & 0.781  & 25.89  & 0.830  & 36.30  & 0.977  & 28.40  & 0.866 & 33.10 & 0.938 \\
          & NR    & 31.62  & 0.924  & 29.73  & 0.868  & 18.41  & 0.627  & 29.17  & 0.911  & 27.23  & 0.833 & 33.39 & 0.941 \\
          & DPL   & 28.32  & 0.888  & 28.48  & 0.849  & 23.93  & 0.796  & 35.95  & 0.967  & 29.17  & 0.875 & 33.87 & 0.945 \\
\cmidrule{2-14}          & CLGID & 31.74  & 0.923  & 32.18  & 0.912  & 28.59  & 0.889  & 37.37  & 0.982  & \underline{\textbf{32.47}}  & \underline{\textbf{0.927}} & 34.03 & \underline{\textbf{0.948}} \\
          & $\text{ CLGID}^{\dag}$ & 31.64  & 0.922  & 32.05  & 0.908  & 28.34  & 0.886  & 37.14  & 0.981  & 32.29  & 0.924 & \underline{\textbf{34.05}}  & 0.947 \\
    \midrule[\heavyrulewidth]
    \multirow{2}[2]{*}{Training Sequence} & \multirow{2}[2]{*}{Methods} & \multicolumn{2}{c}{Rain1400} & \multicolumn{2}{c}{Rain1200M} & \multicolumn{2}{c}{Rain100H} & \multicolumn{2}{c}{Rain100L} & \multicolumn{2}{c}{Avg Memory} & \multicolumn{2}{c}{SPA-data} \\
    \cmidrule{3-14}          &       & PSNR  & SSIM  & PSNR  & SSIM  & PSNR  & SSIM  & PSNR  & SSIM  & PSNR  & SSIM  & PSNR  & SSIM \\
    \midrule
    \multirow{7}[1]{*}{1400-100L-1200M-100H} & \textcolor{black!20}{Individual} & \textcolor{black!20}{32.01}  & \textcolor{black!20}{0.929}  & \textcolor{black!20}{38.33}  & \textcolor{black!20}{0.985}  & \textcolor{black!20}{32.80}  & \textcolor{black!20}{0.925}  & \textcolor{black!20}{29.87}  & \textcolor{black!20}{0.914}  & \textcolor{black!20}{33.25}  & \textcolor{black!20}{0.938}  & \textendash & \textendash \\
          & SF    & 28.72  & 0.880  & 38.15  & 0.983  & 27.13  & 0.841  & 30.22  & 0.920  & 31.06  & 0.906 & 32.60 & 0.929 \\
          & PIGWM & 29.41  & 0.894  & 34.91  & 0.970  & 27.54  & 0.859  & 28.05  & 0.877  & 29.98  & 0.900 & 33.59 & 0.942 \\
          & NR    & 31.59  & 0.924  & 31.78  & 0.944  & 30.27  & 0.876  & 19.70  & 0.676  & 28.34  & 0.855 & 33.35 & 0.945 \\
          & DPL   & 23.20  & 0.770  & 25.10  & 0.834  & 20.67  & 0.683  & 11.70  & 0.387  & 20.17  & 0.669 & 33.26 & 0.940 \\
\cmidrule{2-14}          & CLGID & 31.62  & 0.923  & 37.13  & 0.981  & 32.19  & 0.915  & 29.17  & 0.902  & 32.53  & \underline{\textbf{0.930}} & \underline{\textbf{34.34}} & \underline{\textbf{0.952}}  \\
          & $\text{ CLGID}^{\dag}$ & 31.74  & 0.924  & 37.03  & 0.981  & 32.30  & 0.916  & 29.14  & 0.900  & \underline{\textbf{32.55}}  & \underline{\textbf{0.930}} & 34.10 & 0.947  \\
    \midrule[\heavyrulewidth]
    \multirow{2}[2]{*}{Training Sequence} & \multirow{2}[2]{*}{Methods} & \multicolumn{2}{c}{Rain1400} & \multicolumn{2}{c}{Rain1200M} & \multicolumn{2}{c}{Rain100H} & \multicolumn{2}{c}{Rain100L} & \multicolumn{2}{c}{Avg Memory} & \multicolumn{2}{c}{SPA-data} \\
    \cmidrule{3-14}          &       & PSNR  & SSIM  & PSNR  & SSIM  & PSNR  & SSIM  & PSNR  & SSIM  & PSNR  & SSIM  & PSNR  & SSIM \\
    \midrule
    \multirow{7}[1]{*}{100L-100H-1400-1200M} & \textcolor{black!20}{Individual} & \textcolor{black!20}{38.33}  & \textcolor{black!20}{0.985}  & \textcolor{black!20}{29.87}  & \textcolor{black!20}{0.914}  & \textcolor{black!20}{32.01}  & \textcolor{black!20}{0.929}  & \textcolor{black!20}{32.80}  & \textcolor{black!20}{0.925}  & \textcolor{black!20}{33.25}  & \textcolor{black!20}{0.938} & \textendash & \textendash \\
          & SF    & 24.49  & 0.824  & 12.87  & 0.366  & 28.80  & 0.890  & 32.78  & 0.925  & 24.74  & 0.751 & 29.67 & 0.919 \\
          & PIGWM & 26.16  & 0.888  & 14.46  & 0.468  & 29.03  & 0.899  & 31.66  & 0.904  & 25.33  & 0.790 & 30.96 & 0.930 \\
          & NR    & 36.39  & 0.977  & 20.68  & 0.705  & 28.73  & 0.886  & 27.05  & 0.852  & 28.21  & 0.855 & 33.32 & 0.941 \\
          & DPL   & 32.20  & 0.922  & 20.86  & 0.717  & 29.14  & 0.896  & 28.55  & 0.869  & 27.69  & 0.851 & 33.49 & 0.942 \\
\cmidrule{2-14}          & CLGID & 37.79  & 0.983  & 29.23  & 0.903  & 31.50  & 0.921  & 32.12  & 0.912  & \underline{\textbf{32.66}}  & \underline{\textbf{0.930}} & \underline{\textbf{34.24}}  & \underline{\textbf{0.945}}  \\
          & $\text{ CLGID}^{\dag}$ & 37.08  & 0.980  & 28.13  & 0.885  & 31.33  & 0.919  & 32.01  & 0.911  & 32.14  & 0.924 & 34.13 & 0.941 \\
    \midrule[\heavyrulewidth]
    \multirow{2}[2]{*}{Training Sequence} & \multirow{2}[2]{*}{Methods} & \multicolumn{2}{c}{Rain1400} & \multicolumn{2}{c}{Rain1200M} & \multicolumn{2}{c}{Rain100H} & \multicolumn{2}{c}{Rain100L} & \multicolumn{2}{c}{Avg Memory} & \multicolumn{2}{c}{SPA-data} \\
    \cmidrule{3-14}          &       & PSNR  & SSIM  & PSNR  & SSIM  & PSNR  & SSIM  & PSNR  & SSIM  & PSNR  & SSIM  & PSNR  & SSIM \\
    \midrule
    \multirow{7}[1]{*}{100H-100L-1400-1200M} & \textcolor{black!20}{Individual} & \textcolor{black!20}{29.87}  & \textcolor{black!20}{0.914}  & \textcolor{black!20}{38.33}  & \textcolor{black!20}{0.985}  & \textcolor{black!20}{32.01}  & \textcolor{black!20}{0.929}  & \textcolor{black!20}{32.80}  & \textcolor{black!20}{0.925}  & \textcolor{black!20}{33.25}  & \textcolor{black!20}{0.938} & \textendash & \textendash  \\
          & SF    & 12.92  & 0.366  & 24.64  & 0.825  & 28.93  & 0.887  & 32.83  & 0.925  & 24.83  & 0.751 & 30.07 & 0.930  \\
          & PIGWM & 16.96  & 0.597  & 25.90  & 0.859  & 29.18  & 0.908  & 31.64  & 0.906  & 25.92  & 0.818 & 29.66 & 0.908 \\
          & NR    & 27.06  & 0.864  & 36.26  & 0.976  & 28.90  & 0.883  & 27.54  & 0.840  & 29.94  & 0.891 & 32.77 & 0.936 \\
          & DPL   & 24.95  & 0.845  & 32.58  & 0.924  & 28.54  & 0.876  & 28.85  & 0.857  & 28.73  & 0.876 & 32.96 & 0.938 \\
\cmidrule{2-14}          & CLGID & 27.85  & 0.879  & 37.00  & 0.979  & 31.24  & 0.918  & 31.89  & 0.910  & 32.00  & 0.922 & \underline{\textbf{33.81}}  & \underline{\textbf{0.943}} \\
          & $\text{ CLGID}^{\dag}$ & 28.49  & 0.891  & 37.13  & 0.980  & 31.38  & 0.920  & 31.98  & 0.912  & \underline{\textbf{32.25}}  & \underline{\textbf{0.926}} & 33.45 & 0.942  \\
    \bottomrule
    \end{tabular}%
    }
    \vspace{-0.3cm}
  \label{tab:42}%
\end{table*}%

\begin{table*}[htbp]
  \centering
  \caption{Qualitative comparison of memory performance after training on a stream of four datasets in four distinct sequences. All methods utilize MPRNet \cite{zamir2021multi} as the de-raining network. $\text{CLGID}^{\dag}$ represents the accelerated training version of CLGID, using our proposed similarity-based training speedup algorithm. \textcolor{black!20}{Individual} signifies training and testing on each dataset individually, providing a reference for evaluating the memory performance. We evaluate generalization on SPA-data, \textit{\textbf{which has never been seen during training}}. We highlight the best results using \underline{\textbf{such}} formatting.}
  \setlength{\heavyrulewidth}{1.3pt}
  \setlength{\tabcolsep}{0.15cm}
  \renewcommand{\arraystretch}{0.95}
  \resizebox{0.9\linewidth}{!}{
    \begin{tabular}{cccccccccc|cc|cc}
    \toprule[\heavyrulewidth]
    \multirow{2}[2]{*}{Training Sequence} & \multirow{2}[2]{*}{Methods} & \multicolumn{2}{c}{Rain1400} & \multicolumn{2}{c}{Rain1200M} & \multicolumn{2}{c}{Rain100H} & \multicolumn{2}{c}{Rain100L} & \multicolumn{2}{c}{Avg Memory} & \multicolumn{2}{c}{SPA-data} \\
    \cmidrule{3-14}          &       & PSNR  & SSIM  & PSNR  & SSIM  & PSNR  & SSIM  & PSNR  & SSIM  & PSNR  & SSIM  & PSNR  & SSIM \\
    \midrule
    \multirow{6}[1]{*}{1400-1200M-100H-100L} & \textcolor{black!20}{Individual} & \textcolor{black!20}{31.63}  & \textcolor{black!20}{0.922}  & \textcolor{black!20}{31.93}  & \textcolor{black!20}{0.906}  & \textcolor{black!20}{27.28}  & \textcolor{black!20}{0.885}  & \textcolor{black!20}{36.35}  & \textcolor{black!20}{0.979}  & \textcolor{black!20}{31.80}  & \textcolor{black!20}{0.923} & \textendash & \textendash \\
          & SF    & 26.66  & 0.845  & 21.75  & 0.711  & 19.57  & 0.623  & 37.69  & 0.983  & 26.42  & 0.791 & 33.05 & 0.936 \\
          & PIGWM & 27.23  & 0.866  & 22.83  & 0.759  & 19.34  & 0.660  & 35.88  & 0.975  & 26.32  & 0.815 & 33.24 & 0.943  \\
          & NR    & 30.20  & 0.905  & 26.77  & 0.845  & 23.48  & 0.804  & 34.64  & 0.804  & 28.77  & 0.840 & 33.33 & 0.944 \\
\cmidrule{2-14}          & CLGID & 30.83  & 0.912  & 31.27  & 0.900  & 27.46  & 0.869  & 37.37  & 0.981  & \underline{\textbf{31.73}}  & \underline{\textbf{0.916}} & \underline{\textbf{33.56}}  & \underline{\textbf{0.945}} \\
          & $\text{ CLGID}^{\dag}$ & 30.46  & 0.909  & 31.08  & 0.895  & 26.59  & 0.855  & 37.01  & 0.980  & 31.29  & 0.910 & 33.41 & 0.944 \\
    \midrule[\heavyrulewidth]
    \multirow{2}[2]{*}{Training Sequence} & \multirow{2}[2]{*}{Methods} & \multicolumn{2}{c}{Rain1400} & \multicolumn{2}{c}{Rain1200M} & \multicolumn{2}{c}{Rain100H} & \multicolumn{2}{c}{Rain100L} & \multicolumn{2}{c}{Avg Memory} & \multicolumn{2}{c}{SPA-data} \\
    \cmidrule{3-14}          &       & PSNR  & SSIM  & PSNR  & SSIM  & PSNR  & SSIM  & PSNR  & SSIM  & PSNR  & SSIM  & PSNR  & SSIM \\
    \midrule
    \multirow{6}[1]{*}{1400-100L-1200M-100H} & \textcolor{black!20}{Individual} & \textcolor{black!20}{31.63}  & \textcolor{black!20}{0.922}  & \textcolor{black!20}{36.35}  & \textcolor{black!20}{0.979}  & \textcolor{black!20}{31.93}  & \textcolor{black!20}{0.906}  & \textcolor{black!20}{27.28}  & \textcolor{black!20}{0.885}  & \textcolor{black!20}{31.80}  & \textcolor{black!20}{0.923} & \textendash & \textendash \\
          & SF    & 28.33  & 0.876  & 35.98  & 0.974  & 26.76  & 0.844  & 29.12  & 0.898  & 30.05  & 0.898 & 32.89 & 0.939 \\
          & PIGWM & 29.30  & 0.896  & 33.92  & 0.962  & 28.20  & 0.876  & 27.06  & 0.859  & 29.62  & 0.898 & 33.87 & 0.948 \\
          & NR    & 30.27  & 0.906  & 33.46  & 0.960  & 28.02  & 0.872  & 23.52  & 0.817  & 28.82  & 0.889 & 33.62 & 0.949 \\
\cmidrule{2-14}          & CLGID & 30.76  & 0.911  & 37.33  & 0.980  & 31.23  & 0.896  & 28.99  & 0.895  & \underline{\textbf{32.08}}  & \underline{\textbf{0.921}} & \underline{\textbf{34.26}}  & \underline{\textbf{0.951}} \\
          & $\text{ CLGID}^{\dag}$ & 30.18  & 0.904  & 37.22  & 0.979  & 30.76  & 0.891  & 28.47  & 0.884  & 31.66  & 0.915 & 34.09 & 0.949  \\
    \midrule[\heavyrulewidth]
    \multirow{2}[2]{*}{Training Sequence} & \multirow{2}[2]{*}{Methods} & \multicolumn{2}{c}{Rain1400} & \multicolumn{2}{c}{Rain1200M} & \multicolumn{2}{c}{Rain100H} & \multicolumn{2}{c}{Rain100L} & \multicolumn{2}{c}{Avg Memory} & \multicolumn{2}{c}{SPA-data} \\
    \cmidrule{3-14}          &       & PSNR  & SSIM  & PSNR  & SSIM  & PSNR  & SSIM  & PSNR  & SSIM  & PSNR  & SSIM  & PSNR  & SSIM \\
    \midrule
    \multirow{6}[1]{*}{100L-100H-1400-1200M} & \textcolor{black!20}{Individual} & \textcolor{black!20}{36.35}  & \textcolor{black!20}{0.979}  & \textcolor{black!20}{27.28}  & \textcolor{black!20}{0.885}  & \textcolor{black!20}{31.63}  & \textcolor{black!20}{0.922}  & \textcolor{black!20}{31.93}  & \textcolor{black!20}{0.906}  & \textcolor{black!20}{31.80}  & \textcolor{black!20}{0.923} & \textendash & \textendash \\
          & SF    & 24.17  & 0.820  & 12.94  & 0.368  & 28.46  & 0.885  & 32.32  & 0.915  & 24.47  & 0.747 & 27.60 & 0.905 \\
          & PIGWM & 26.38  & 0.873  & 14.06  & 0.413  & 27.63  & 0.878  & 29.34  & 0.873  & 24.35  & 0.759 & 31.63 & 0.935 \\
          & NR    & 31.46  & 0.941  & 16.02  & 0.488  & 29.39  & 0.895  & 27.89  & 0.869  & 26.19  & 0.798 & 31.87 & 0.938 \\
\cmidrule{2-14}          & CLGID & 35.51  & 0.973  & 27.14  & 0.865  & 30.78  & 0.913  & 31.97  & 0.911  & \underline{\textbf{31.26}}  & \underline{\textbf{0.916}} & \underline{\textbf{34.39}}  & \underline{\textbf{0.953}} \\
          & $\text{ CLGID}^{\dag}$ & 34.58  & 0.967  & 25.88  & 0.834  & 29.26  & 0.901  & 31.56  & 0.904  & 30.32  & 0.902 & 34.13 & 0.952 \\
    \midrule[\heavyrulewidth]
   \multirow{2}[2]{*}{Training Sequence} & \multirow{2}[2]{*}{Methods} & \multicolumn{2}{c}{Rain1400} & \multicolumn{2}{c}{Rain1200M} & \multicolumn{2}{c}{Rain100H} & \multicolumn{2}{c}{Rain100L} & \multicolumn{2}{c}{Avg Memory} & \multicolumn{2}{c}{SPA-data} \\
    \cmidrule{3-14}          &       & PSNR  & SSIM  & PSNR  & SSIM  & PSNR  & SSIM  & PSNR  & SSIM  & PSNR  & SSIM  & PSNR  & SSIM \\
    \midrule
    \multirow{6}[1]{*}{100H-100L-1400-1200M} & \textcolor{black!20}{Individual} & \textcolor{black!20}{27.28}  & \textcolor{black!20}{0.885}  & \textcolor{black!20}{36.35}  & \textcolor{black!20}{0.979}  & \textcolor{black!20}{31.63}  & \textcolor{black!20}{0.922}  & \textcolor{black!20}{31.93}  & \textcolor{black!20}{0.906}  & \textcolor{black!20}{31.80}  & \textcolor{black!20}{0.923} & \textendash & \textendash \\
          & SF    & 12.94  & 0.365  & 24.14  & 0.817  & 28.30  & 0.878  & 32.40  & 0.917  & 24.45  & 0.744 & 28.40 & 0.914 \\
          & PIGWM & 13.29  & 0.393  & 25.17  & 0.849  & 26.83  & 0.863  & 28.37  & 0.859  & 23.42  & 0.741 & 32.37 & 0.937 \\
          & NR    & 15.17  & 0.453  & 30.46  & 0.923  & 28.62  & 0.880  & 27.30  & 0.852  & 25.39  & 0.777 & 33.05 & 0.946 \\
\cmidrule{2-14}          & CLGID & 26.07  & 0.841  & 35.41  & 0.973  & 30.73  & 0.913  & 31.94  & 0.910  & \underline{\textbf{31.04}}  & \underline{\textbf{0.909}} & \underline{\textbf{34.60}}  & \underline{\textbf{0.956}} \\
          & $\text{ CLGID}^{\dag}$ & 25.98  & 0.843  & 34.98  & 0.969  & 30.14  & 0.911  & 31.70  & 0.908  & 30.70  & 0.908 & 34.22 & 0.951 \\
    \bottomrule
    \end{tabular}%
    }
    \vspace{-0.3cm}
  \label{tab:43}%
\end{table*}%

\begin{table*}[htbp]
  \centering
  \caption{Qualitative comparison of memory performance after training on a stream of six datasets in 1400-1200M-100H-100L-1200L-1200H sequence. $\text{CLGID}^{\dag}$ represents the accelerated training version of CLGID, using our proposed similarity-based training speedup algorithm. \textcolor{black!20}{Individual} signifies training and testing on each dataset individually, providing a reference for evaluating the memory performance.  We evaluate generalization on SPA-data, \textit{\textbf{which has never been seen during training}}. We highlight the best results using \underline{\textbf{such}} formatting.}
  \setlength{\heavyrulewidth}{1.3pt}
  \setlength{\tabcolsep}{0.15cm}
  \resizebox{0.95\linewidth}{!}{%
    \begin{tabular}{cccccccccccccc|cc|cc}
    \toprule
    \multirow{2}[2]{*}{Network} & \multirow{2}[2]{*}{Methods} & \multicolumn{2}{c}{Rain1400} & \multicolumn{2}{c}{Rain1200M} & \multicolumn{2}{c}{Rain100H} & \multicolumn{2}{c}{Rain100L} & \multicolumn{2}{c}{Rain1200L} & \multicolumn{2}{c}{Rain1200H} & \multicolumn{2}{c}{Avg Memory} & \multicolumn{2}{c}{SPA-data} \\
\cmidrule{3-18}          &       & PSNR  & SSIM  & PSNR  & SSIM  & PSNR  & SSIM  & PSNR  & SSIM  & PSNR  & SSIM  & PSNR  & SSIM  & PSNR  & SSIM & PSNR  & SSIM \\
    \midrule[\heavyrulewidth]
    \multirow{7}[1]{*}{MFDNet} & \textcolor{black!20}{Individual} & \textcolor{black!20}{31.79}  & \textcolor{black!20}{0.920}  & \textcolor{black!20}{32.24}  & \textcolor{black!20}{0.920}  & \textcolor{black!20}{27.70}  & \textcolor{black!20}{0.886}  & \textcolor{black!20}{36.16}  & \textcolor{black!20}{0.978}  & \textcolor{black!20}{36.22}  & \textcolor{black!20}{0.958}  & \textcolor{black!20}{29.91}  & \textcolor{black!20}{0.889}  & \textcolor{black!20}{32.34}  & \textcolor{black!20}{0.925}  & \textendash & \textendash \\
          & SF    & 26.88  & 0.877  & 30.78  & 0.914  & 13.73  & 0.384  & 23.16  & 0.804  & 27.16  & 0.900  & 30.83  & 0.897  & 25.42  & 0.796 & 31.76 & 0.929 \\
          & PIGWM & 29.09  & 0.899  & 30.38  & 0.900  & 14.72  & 0.446  & 25.59  & 0.850  & 31.34  & 0.921  & 29.83  & 0.878  & 26.83  & 0.816 & 32.39 & 0.940 \\
          & NR    & 31.16  & 0.917  & 30.34  & 0.899  & 15.41  & 0.493  & 29.53  & 0.916  & 34.87  & 0.948  & 26.30  & 0.827  & 27.94  & 0.833 & 32.59 & 0.943 \\
          & DPL   & 30.51  & 0.886  & 29.74  & 0.885  & 22.33  & 0.782  & 28.88  & 0.902  & 30.30  & 0.903  & 26.77  & 0.827  & 28.09  & 0.864 & 32.08 & 0.938\\
\cmidrule{2-18}          & CLGID & 30.73  & 0.908  & 31.07  & 0.913  & 27.30  & 0.865  & 36.49  & 0.978  & 35.97  & 0.954  & 30.40  & 0.889  & \underline{\textbf{31.99}}  & \underline{\textbf{0.918}} & \underline{\textbf{34.36}}  & \underline{\textbf{0.952}}  \\
          & $\text{ CLGID}^{\dag}$ & 30.99  & 0.912  & 31.51  & 0.916  & 26.60  & 0.842  & 36.20  & 0.976  & 36.03  & 0.957  & 30.37  & 0.887  & 31.95  & 0.915 & 34.23 & 0.950 \\
    \midrule[\heavyrulewidth]
    \multirow{7}[1]{*}{Restormer} & \textcolor{black!20}{Individual} & \textcolor{black!20}{32.01}  & \textcolor{black!20}{0.929}  & \textcolor{black!20}{32.80}  & \textcolor{black!20}{0.925}  & \textcolor{black!20}{29.87}  & \textcolor{black!20}{0.914}  & \textcolor{black!20}{38.33}  & \textcolor{black!20}{0.985}  & \textcolor{black!20}{36.54}  & \textcolor{black!20}{0.960}  & \textcolor{black!20}{30.95}  & \textcolor{black!20}{0.898}  & \textcolor{black!20}{33.42}  & \textcolor{black!20}{0.935} & \textendash & \textendash \\
          & SF    & 27.38  & 0.878  & 31.18  & 0.918  & 13.42  & 0.374  & 23.67  & 0.812  & 27.84  & 0.912  & 31.17  & 0.900  & 25.78  & 0.799 & 30.14 & 0.915 \\
          & PIGWM & 27.98  & 0.886  & 30.20  & 0.891  & 14.45  & 0.434  & 25.00  & 0.837  & 30.53  & 0.899  & 30.19  & 0.880  & 26.39  & 0.805 & 32.24 & 0.929 \\
          & NR    & 31.27  & 0.919  & 30.55  & 0.888  & 13.95  & 0.435  & 27.63  & 0.888  & 35.22  & 0.949  & 25.97  & 0.825  & 27.43  & 0.817 & 33.06 & 0.941 \\
          & DPL   & 29.95  & 0.907  & 30.43  & 0.882  & 15.73  & 0.554  & 25.85  & 0.867  & 33.70  & 0.933  & 25.56  & 0.826  & 26.87  & 0.828 & 32.28 & 0.930 \\
\cmidrule{2-18}          & CLGID & 31.70  & 0.923  & 32.02  & 0.915  & 28.26  & 0.884  & 37.26  & 0.981  & 36.20  & 0.957  & 30.51  & 0.885  & \underline{\textbf{32.66}}  & \underline{\textbf{0.924}} & 34.34 & \underline{\textbf{0.951}}  \\
          & $\text{ CLGID}^{\dag}$ & 31.63  & 0.922  & 31.93  & 0.913  & 28.18  & 0.882  & 37.16  & 0.981  & 36.09  & 0.956  & 30.49  & 0.888  & 32.58  & \underline{\textbf{0.924}} & \underline{\textbf{34.35}} & 0.950 \\
    \midrule[\heavyrulewidth]
    \multirow{6}[1]{*}{MPRNet} & \textcolor{black!20}{Individual} & \textcolor{black!20}{31.63}  & \textcolor{black!20}{0.922}  & \textcolor{black!20}{31.93}  & \textcolor{black!20}{0.906}  & \textcolor{black!20}{27.28}  & \textcolor{black!20}{0.885}  & \textcolor{black!20}{36.35}  & \textcolor{black!20}{0.979}  & \textcolor{black!20}{35.02}  & \textcolor{black!20}{0.942}  & \textcolor{black!20}{28.08}  & \textcolor{black!20}{0.856}  & \textcolor{black!20}{31.72}  & \textcolor{black!20}{0.915} & \textendash & \textendash \\
          & SF    & 26.24  & 0.867  & 30.54  & 0.907  & 13.59  & 0.379  & 22.48  & 0.783  & 25.91  & 0.877  & 30.57  & 0.888  & 24.89  & 0.784 & 28.54 & 0.897 \\
          & PIGWM & 29.04  & 0.904  & 30.60  & 0.899  & 16.14  & 0.526  & 26.85  & 0.884  & 32.61  & 0.936  & 29.66  & 0.871  & 27.48  & 0.837 & 32.13 & 0.937 \\
          & NR    & 30.68  & 0.909  & 30.02  & 0.896  & 14.34  & 0.457  & 29.30  & 0.916  & 34.62  & 0.945  & 26.36  & 0.844  & 27.55  & 0.828 & 32.34 & 0.939 \\
\cmidrule{2-18}          & CLGID & 30.68  & 0.907 & 31.15  & 0.906  & 26.91  & 0.858  & 35.69  & 0.974  & 35.23  & 0.944  & 30.22  & 0.882  & \underline{\textbf{31.65}}  & \underline{\textbf{0.912}} & 34.30 & 0.953 \\
          & $\text{CLGID}^{\dag}$ & 29.76  & 0.894  & 31.17  & 0.907  & 26.47  & 0.849  & 35.22  & 0.972  & 35.15  & 0.942  & 30.19  & 0.881  & 31.33  & 0.908 & \underline{\textbf{34.34}}  & \underline{\textbf{0.955}} \\
    \bottomrule
    \end{tabular}%
    }
  \label{tab:61}%
\end{table*}%

\textbf{Comparison methods.}
We conduct comparative experiments on memory and generalization ability, utilizing a baseline (denoted by ``SF''), along with three related state-of-the-art methods: PIGWM \cite{zhou2021image}, NR \cite{xiao2021improving}, and DPL \cite{liu2023dual}, leveraging their publicly released codes. SF entails the sequential fine-tuning of the de-raining network on each new incoming dataset. PIGWM \cite{zhou2021image} employs the parameter importance guided weights modification method to enable the de-raining networks to learn from a sequence of datasets. 
NR \cite{xiao2021improving} explores a neural reorganization method to ensure the accumulation of de-raining knowledge. DPL \cite{liu2023dual} utilizes a dual prompt learning scheme designed to handle diverse types of rain streaks within a single de-raining transformer. Additionally, we introduce individual training (denoted by ``Individual''), which trains and tests on each single dataset within a stream of datasets, as a reference to evaluate the memory performance of several methods.

\textbf{Evaluation Metrics.}
Memory and generalization performance are evaluated in terms of Peak Signal-to-Noise Ratio (PSNR), and Structural Similarity (SSIM) metrics. We compute PSNR and SSIM metrics over RGB channels for color images.

\textbf{Implementation Details.}
Our approach is implemented in PyTorch using NVIDIA 3090 GPUs. To ensure a fair comparison, we set the patch size of all methods to 64, including baseline, SOTA methods, and our approach. We conduct experiments on three representative de-raining networks: MFDNet \cite{wang2023multi}, Restormer \cite{zamir2022restormer}, and MPRNet \cite{zamir2021multi}. The training settings of the de-raining networks remain consistent with their publicly released code, including batch size, training epochs, iterations, optimizer, scheduler, etc. Note that DPL is designed for transformer-based de-raining networks; therefore, we only evaluate DPL on MFDNet and Restormer. The hyper-parameter of our framework, $\lambda$, which balances the interleave loss and the consistency loss, is set to 1.
{The $\hat{T}$ in similarity-based selective GAN training is set to 0.4.}

\begin{figure}[htbp]
\begin{center}
\includegraphics[width=\linewidth]{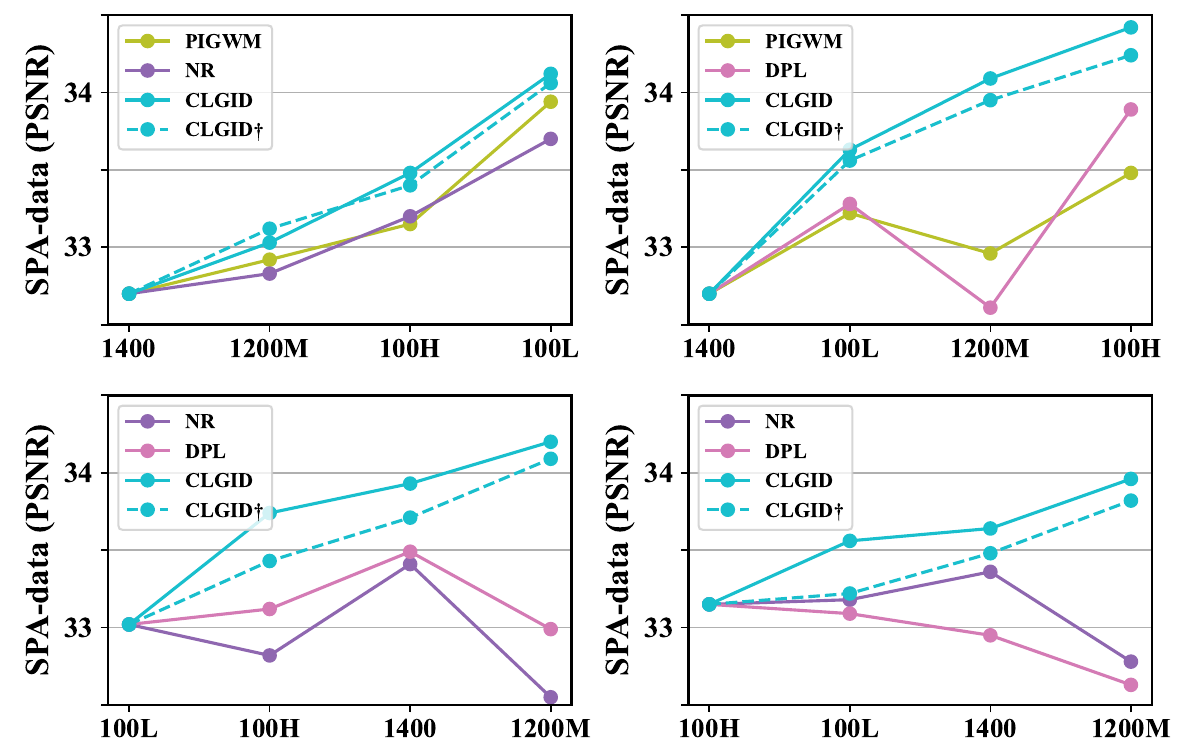}
\end{center}
\vspace{-0.2cm}
\caption{Generalization performance variance on unseen SPA-data during training on on a stream of four datasets across four sequences using MFDNet \cite{wang2023multi}. We showcase the top four methods on generalization results.}
\label{fig4}
\end{figure}

\begin{figure}[htbp]
\begin{center}
\includegraphics[width=\linewidth]{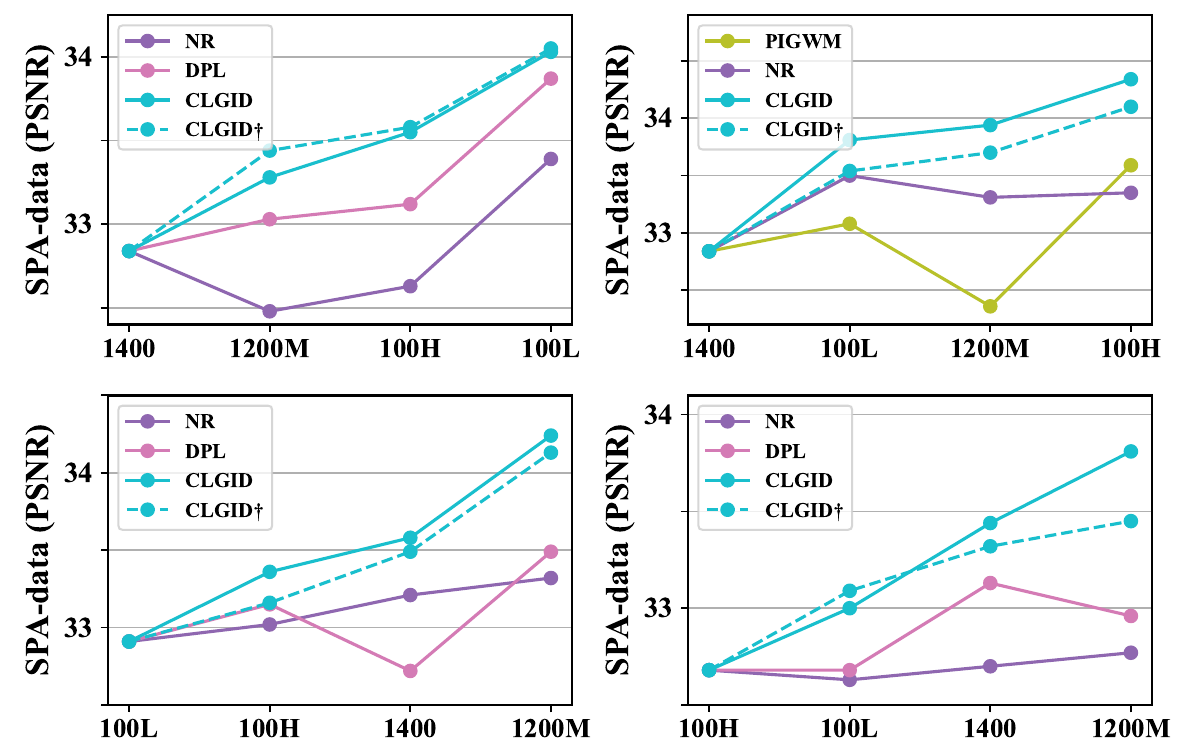}
\end{center}
\caption{Generalization performance variance on unseen SPA-data during training on on a stream of four datasets across four sequences using Restormer \cite{zamir2022restormer}. We showcase the top four methods on generalization.}
\label{fig5}
\end{figure}

\begin{figure}[htbp]
\begin{center}
\includegraphics[width=\linewidth]{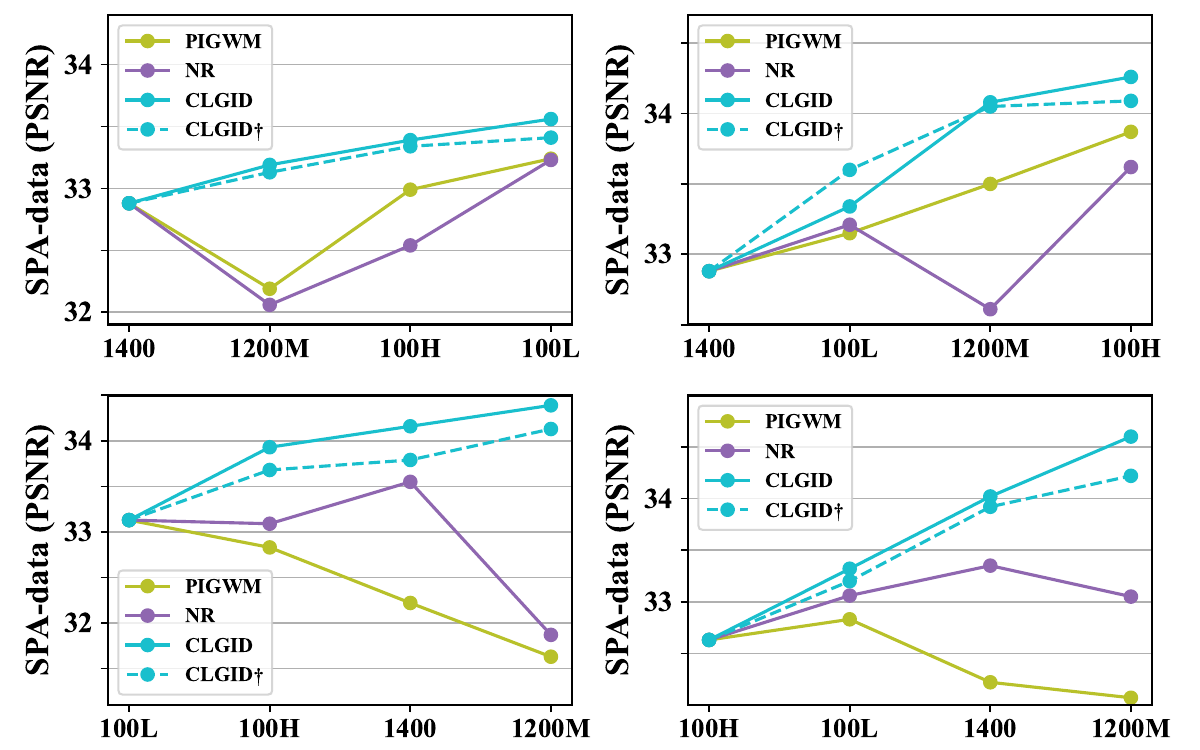}
\end{center}
\caption{Generalization performance variance on unseen SPA-data during training on a stream of four datasets across four sequences using MPRNet \cite{zamir2021multi}. We showcase the top four methods on generalization.}
\label{fig6}
\end{figure}

\begin{figure*}[h]
\begin{center}
\includegraphics[width=\linewidth]{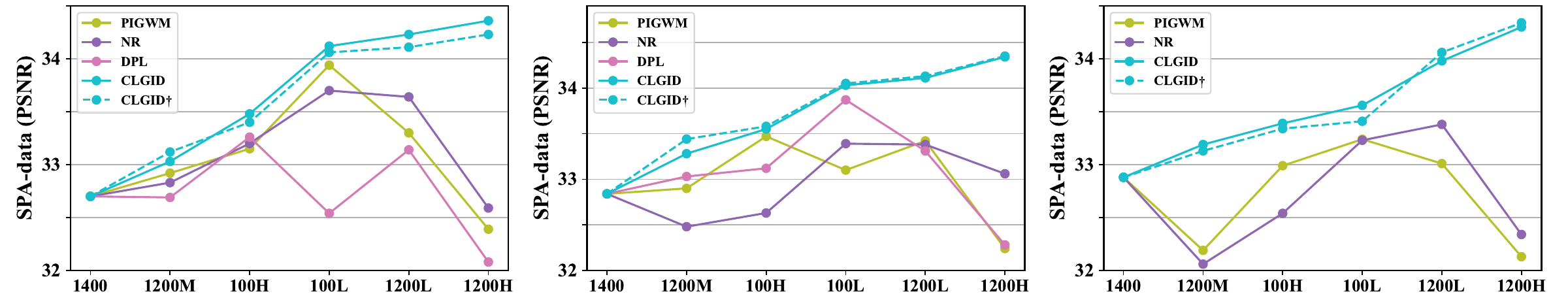}
\end{center}
\caption{Generalization performance variance on unseen SPA-data during training on a stream of six datasets. We showcase the top five methods on generalization. The three charts from left to right respectively represent using MFDNet \cite{wang2023multi}, Restormer \cite{zamir2022restormer}, and MPRNet \cite{zamir2021multi}.}
\vspace{-0.3cm}
\label{fig7}
\end{figure*}

\subsection{Results on Benchmark Datasets}
{To validate the efficacy of our approach, we conduct twelve experiments using three representative de-raining networks, i.e., MFDNet~\cite{wang2023multi}, Restormer~\cite{zamir2022restormer}, and MPRNet~\cite{zamir2021multi}, on four representative training sequences composed of four de-raining datasets. These sequences are deliberately designed to reflect variations in rain type diversity, intensity, and dataset complexity, enabling a systematic evaluation of the proposed method’s robustness to training order in continual learning scenarios.}
After training, we evaluate both memory performance on the seen datasets and generalization performance on the unseen real-world SPA-data~\cite{wang2019spatial}. 
As shown in Tab.~\ref{tab:41}, \ref{tab:42}, and \ref{tab:43}, sequential fine-tuning (SF) suffers from severe forgetting, leading to poor generalization. Compared with state-of-the-art methods such as PIGWM, NR, and DPL, our CLGID achieves substantial improvements in memory retention and matches the performance of the Individual baseline, which represents an upper bound by training separately on each dataset. Regarding generalization, CLGID consistently outperforms all competitors across all settings.

Furthermore, we analyze the variation in generalization performance as more datasets are introduced, as illustrated in Fig.~\ref{fig4}, \ref{fig5}, and \ref{fig6}. While other methods show limited or unstable improvement due to memory saturation, our method exhibits consistent gains in generalization across all experiments, demonstrating its strong knowledge accumulation ability and robustness to variations in training sequences.

In addition to quantitative results, we also provide qualitative assessments of various methods on SPA-data \cite{wang2019spatial} and Real-internet \cite{wang2019spatial}, after training on the 1400-1200M-100H-100L sequence, as shown in Fig.~\ref{fig8}. We can observe that other methods struggle to eliminate heavy rain streaks and those resembling the background's texture. Some artifacts persist in the de-raining outcomes, and the background details appear blurred. In contrast, our CLGID yields the most visually appealing results.

\begin{table}[htbp]
  \centering
  \caption{{Memory and generalization performance on two alternative six-dataset sequences using MPRNet~\cite{zamir2021multi}.}}
  \setlength{\heavyrulewidth}{1.3pt}
  \setlength{\tabcolsep}{0.15cm}
  \resizebox{\linewidth}{!}{%
    \begin{tabular}{cccccc}
    \toprule
    \multirow{2}[2]{*}{Training Sequence} & \multirow{2}[2]{*}{Methods} & \multicolumn{2}{c}{Avg Memory} & \multicolumn{2}{c}{SPA-data} \\
\cmidrule{3-6}          &       & PSNR  & SSIM  & PSNR  & SSIM \\
    \midrule
    \multirow{6}[3]{*}{1400-100L-1200M-100H-1200H-1200L} & Individual & 31.72  & 0.915 &  \textendash  &  \textendash \\
          & SF    & 28.16  & 0.851 & 30.01  & 0.923  \\
          & PIGWM & 29.01  & 0.875 & 33.24  & 0.940  \\
          & NR    & 27.94  & 0.834 & 33.60  & 0.947  \\
\cmidrule{2-6}          & CLGID & \underline{\textbf{31.92}}  & \underline{\textbf{0.919}} & \underline{\textbf{34.29}}  & \underline{\textbf{0.955}}  \\
          & $\text{CLGID}^{\dag}$ & 31.51  & 0.913 & 34.21  & 0.951  \\
    \midrule
    \multirow{6}[3]{*}{100L-100H-1400-1200M-1200L-1200H} & Individual & 31.72  & 0.915 &  \textendash  &  \textendash \\
          & SF    & 22.10  & 0.705 & 28.12  & 0.907  \\
          & PIGWM & 24.51  & 0.778 & 31.11  & 0.920  \\
          & NR    & 26.48  & 0.813 & 30.23  & 0.914  \\
\cmidrule{2-6}          & CLGID & \underline{\textbf{31.12}}  & \underline{\textbf{0.914}} & \underline{\textbf{34.45}}  & \underline{\textbf{0.957}}  \\
          &$\text{CLGID}^{\dag}$ & 30.60  & 0.911 & 34.33  & 0.955  \\
    \bottomrule
    \end{tabular}%
    }
  \vspace{-0.3cm}
  \label{tab:six_more}%
\end{table}%

\begin{figure*}[htbp]
\begin{center}
\includegraphics[width=\linewidth]{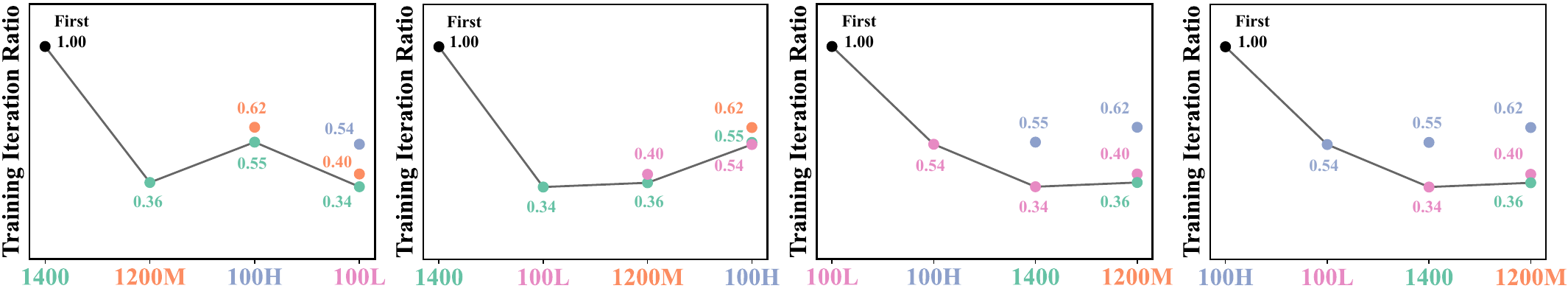}
\end{center}
\caption{Training iteration ratios of $\text{CLGID}^\dag$ compared with CLGID training on a stream of four datasets across four sequences. Each data point in the plots represents the similarity calculated between the current dataset and the previously learned dataset with the same color.}
\vspace{-0.3cm}
\label{speed}
\end{figure*}

\begin{figure*}[htbp]
\begin{center}
\includegraphics[width=\linewidth]{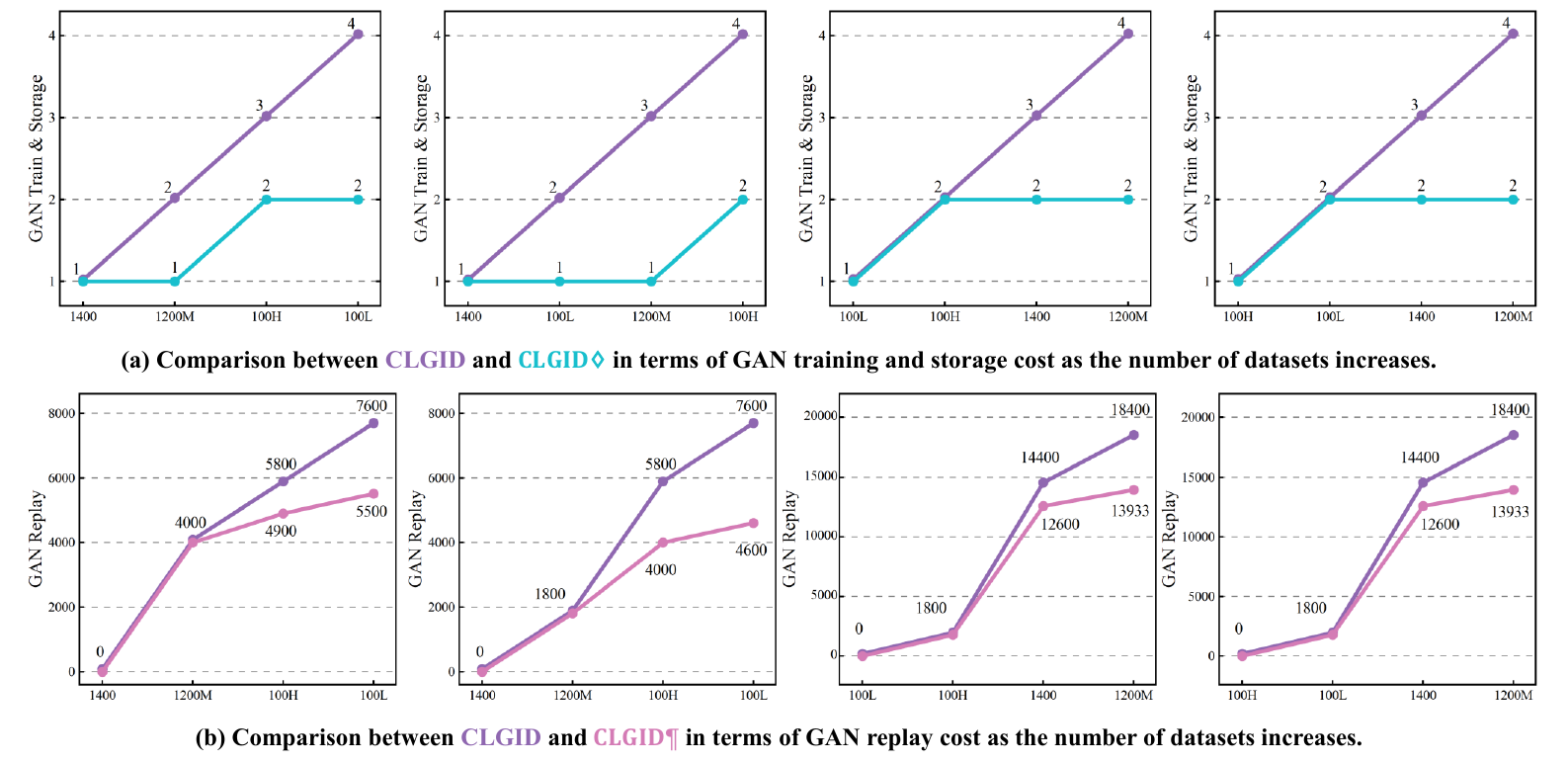}
\end{center}
\vspace{-0.2cm}
\caption{{Comparison of GAN training, storage, and replay costs between CLGID and its variants, $\text{CLGID}^{\lozenge}$ and CLGID\textsuperscript{\textparagraph}, on a four-dataset stream across four sequences. The y-axis represents the number of times.}}
\vspace{-0.3cm}
\label{gan_scale}
\end{figure*}

\subsection{Extension to More Datasets}
To further evaluate the memory retention and generalization capability of CLGID on more datasets, we extend the training sequence to include six datasets: 1400-1200M-100H-100L-1200L-1200H. The corresponding results on memory and generalization performance after training, as well as the generalization trend during training, are presented in Tab.~\ref{tab:61} and Fig.~\ref{fig7}. CLGID consistently outperforms existing methods in both metrics, demonstrating strong resistance to catastrophic forgetting. Compared with its performance under the four-dataset setting (1400-1200M-100H-100L), CLGID not only preserves the knowledge of previously seen datasets but also achieves further improvements in generalization to unseen real-world images. In contrast, competing methods suffer significant drops in generalization performance as more datasets are introduced, highlighting their limited memory capacity and inability to accumulate knowledge effectively.

To provide a comprehensive evaluation, we select one representative six-dataset sequence for the main analysis. This sequence integrates diverse data sources and rain patterns, introducing substantial distribution shifts that serve as a strong testbed for evaluating continual knowledge accumulation. While exhaustive enumeration of all possible permutations is infeasible, we report results on two alternative six-dataset sequences using MPRNet: 1400-100L-1200M-100H-1200H-1200L and 100L-100H-1400-1200M-1200L-1200H, as shown in Table~\ref{tab:six_more}. Our method maintains superior performance across these variations, reinforcing the robustness of our framework.

\subsection{Ablation Study}
\textbf{Validation on the training speedup algorithm.}
$\text{CLGID}^{\dag}$ represents the version of CLGID that utilizes the proposed similarity-based training speedup algorithm. Fig. \ref{speed} showcase the ratio of total training iterations of $\text{ CLGID}^{\dag}$ compared with CLGID. Training on a stream of four datasets in four different sequences, $\text{ CLGID}^{\dag}$ achieves an average reduction of 44\% in total training iterations and 42\% in total training time. 
The varying reductions in total training iterations and training time are attributable to the time required to calculate similarity.
Furthermore, as shown in Tab. \ref{tab:41}-\ref{tab:43} and Fig. \ref{fig4}-\ref{fig6}, we can observe that $\text{ CLGID}^{\dag}$ achieves comparable memory and generalization performance compared to CLGID.
The above observations underscore the algorithm's efficacy to shorten the training time without compromising the de-raining network’s memory generalization ability, resulting in low training costs.

\begin{table}[htbp]
  \centering
  \caption{{Memory and generalization performance of CLGID and its variants $\text{CLGID}^{\lozenge}$ and CLGID\textsuperscript{\textparagraph} across four four-dataset sequences using MPRNet.}}
  \resizebox{\linewidth}{!}{
    \begin{tabular}{cccccc}
    \toprule
    \multirow{2}[2]{*}{Training Sequence} & \multirow{2}[2]{*}{Methods} & \multicolumn{2}{c}{Avg Memory} & \multicolumn{2}{c}{SPA-data} \\
\cmidrule{3-6}          &       & PSNR  & SSIM  & PSNR  & SSIM \\
    \midrule
    \multirow{3}[3]{*}{1400-1200M-100H-100L} & CLGID & 31.73  & 0.916  & 33.56  & 0.945  \\
\cmidrule{2-6}          & $\text{CLGID}^{\lozenge}$ & 31.11  & 0.908  & 33.40  & 0.943  \\
\cmidrule{2-6}          & CLGID\textsuperscript{\textparagraph} & 31.70  & 0.915  & 33.50  & 0.945  \\
    \midrule
    \multirow{3}[3]{*}{1400-100L-1200M-100H} & CLGID & 32.08  & 0.921  & 34.26  & 0.951  \\
\cmidrule{2-6}          & $\text{CLGID}^{\lozenge}$ & 31.44  & 0.909  & 34.03  & 0.948  \\
\cmidrule{2-6}          & CLGID\textsuperscript{\textparagraph} & 32.08  & 0.920  & 34.27  & 0.951  \\
    \midrule
    \multirow{3}[3]{*}{100L-100H-1400-1200M} & CLGID & 31.26  & 0.916  & 34.39  & 0.953  \\
\cmidrule{2-6}          & $\text{CLGID}^{\lozenge}$ & 30.20  & 0.900  & 34.21  & 0.951  \\
\cmidrule{2-6}          & CLGID\textsuperscript{\textparagraph} & 31.25  & 0.917  & 34.36  & 0.953  \\
    \midrule
    \multirow{3}[3]{*}{100H-100L-1400-1200M} & CLGID & 31.04  & 0.909  & 34.60  & 0.956  \\
\cmidrule{2-6}          & $\text{CLGID}^{\lozenge}$ & 30.49  & 0.905  & 34.20  & 0.952  \\
\cmidrule{2-6}          & CLGID\textsuperscript{\textparagraph} & 31.02  & 0.908  & 34.61  & 0.955  \\
    \bottomrule
    \end{tabular}%
    }
  \label{tab:GAN}%
\end{table}%

\begin{figure*}[t]
\begin{center}
\includegraphics[width=\linewidth]{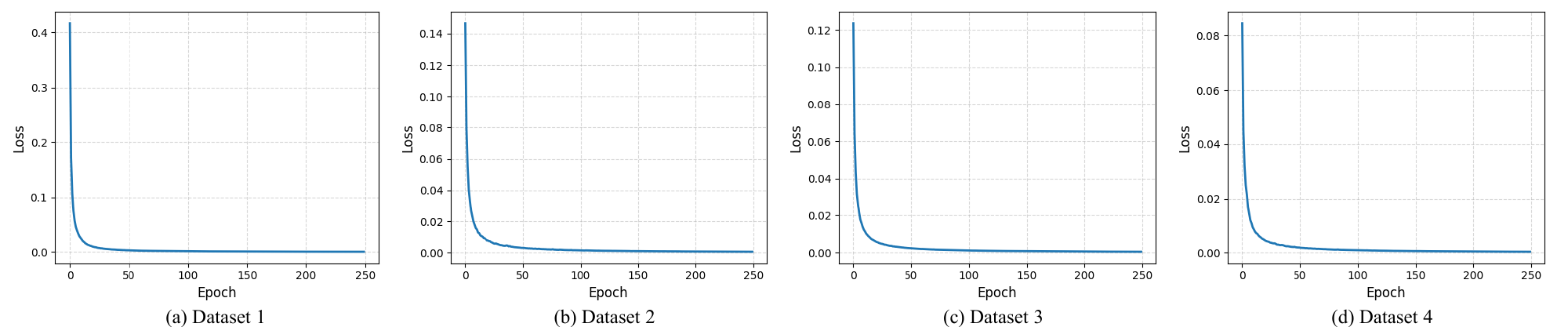}
\end{center}
\caption{{Training loss curve under the sequence 1400-1200M-100H-100 using MPRNet.}}
\vspace{-0.4cm}
\label{Loss}
\end{figure*}

{\textbf{Validation of GAN scalability.}
$\text{CLGID}^{\lozenge}$ and CLGID\textsuperscript{\textparagraph} denote variants of the proposed CLGID framework that incorporate the similarity-based selective GAN training and GAN-replayed data reuse mechanisms, respectively. We validate their effectiveness using MPRNet on four distinct four-dataset sequences. The experimental results are presented in Fig.~\ref{gan_scale} and Tab.~\ref{tab:GAN}.
It is evident that both $\text{CLGID}^{\lozenge}$ and CLGID\textsuperscript{\textparagraph} significantly enhance the scalability of the framework compared with the original CLGID. Specifically, as the number of datasets increases, $\text{CLGID}^{\lozenge}$ achieves an average reduction of over 50\% in GAN training FLOPs, training time, and parameter storage. Meanwhile, CLGID\textsuperscript{\textparagraph} reduces the number of GAN inference calls required for replay by an average of 26.9\%, thereby decreasing the overall replay-related FLOPs and time consumption.
In terms of performance, $\text{CLGID}^{\lozenge}$ exhibits a slight degradation in memory and generalization, but it still significantly outperforms all other SOTA methods. 
CLGID\textsuperscript{\textparagraph} maintains nearly the same level of memory and generalization as the original CLGID, as the replay data reuse strategy does not incur meaningful information loss during training.}

\textbf{Validation on each framework component.}
To validate the effectiveness of each loss function, we train our method under circumstances that remove replay loss and consistency loss in a successive manner. The results are shown in Table~\ref{tab:loss}, and it is evident that each loss contributes to the promotion of memory performance on training datasets and generalization performance on the real-world dataset. The proposed CLGID achieves the best performance.

\begin{table}[t]
  \centering
  \caption{Analysis of the efficacy of each CLGID component in training on the 1400-100L-1200M-100H sequence using MPRNet \cite{zamir2021multi}.}
  \setlength{\tabcolsep}{0.15cm}
  \renewcommand{\arraystretch}{1.1}
  \resizebox{\linewidth}{!}{
    \begin{tabular}{ccccc}
    \toprule
          & w/o both & w/o $\mathcal{L}_{\text{replay}}$ & w/o $\mathcal{L}_{\text{consist}}$ & w both \\
    \midrule
    Rain1400 & 28.33 / 0.876 & 29.64 / 0.891 & 29.87 / 0.900 & 30.76 / 0.911 \\
    Rain100L & 35.98 / 0.974 & 36.88 / 0.974 & 36.26 / 0.970 & 37.33 / 0.980 \\
    Rain1200M & 26.76 / 0.844 & 27.07 / 0.846 & 29.18 / 0.874 & 31.23 / 0.896 \\
    Rain100H & 29.12 / 0.898 & 28.58 / 0.892 & 28.93 / 0.894 & 28.99 / 0.895 \\
    \midrule
    SPA-data & 32.89 / 0.939 & 33.68 / 0.943 & 33.86 / 0.947 & 34.26 / 0.951 \\
    \bottomrule
    \end{tabular}%
    }
  \label{tab:loss}%
\end{table}%

\begin{table}[t]
  \centering
  \caption{Analysis of hyper-parameter $\lambda$ in training on the 1400-100L-1200M-100H sequence using MPRNet \cite{zamir2021multi}.}
  \setlength{\tabcolsep}{0.15cm}
  \renewcommand{\arraystretch}{1.1}
  \resizebox{\linewidth}{!}{
    \begin{tabular}{cccccc}
    \toprule
          & $\lambda=0.1$   & $\lambda=0.5$   & $\lambda=1.0$     & $\lambda=2.0$     & $\lambda=5.0$ \\
    \midrule
    Rain1400 & 29.91 / 0.901 & 29.80 / 0.902 & 30.76 /0.911 & 29.75 / 0.899 & 29.82 / 0.897 \\
    Rain100L & 37.08 / 0.975 & 37.19 / 0.976 & 37.33 / 0.980 & 37.10 / 0.976 & 37.07 / 0.973 \\
    Rain1200M & 28.35 / 0.848 & 29.41 / 0.860 & 31.23 / 0.896 & 30.26 / 0.876 & 28.47 / 0.861 \\
    Rain100H & 28.62 / 0.892 & 28.79 / 0.895 & 28.99 / 0.895 & 28.74 / 0.892 & 28.73 / 0.891 \\
    \midrule
    SPA-data & 34.03 / 0.947 & 34.07 / 0.948 & 34.26 / 0.951 & 33.84 / 0.943 & 33.70 / 0.942 \\
    \bottomrule
    \end{tabular}%
    }
  \label{tab:hyper}%
  \vspace{-0.3cm}
\end{table}%

\begin{table}[t]
  \centering
  \caption{{Stability analysis of the framework. Mean and standard deviation of performance over 10 runs are reported.}}
  \setlength{\tabcolsep}{0.15cm}
  \renewcommand{\arraystretch}{1.1}
  \resizebox{\linewidth}{!}{
    \begin{tabular}{cccccc}
    \toprule
    \multirow{2}[2]{*}{Training Sequence} & \multirow{2}[2]{*}{Methods} & \multicolumn{2}{c}{Avg Memory} & \multicolumn{2}{c}{SPA-data} \\
\cmidrule{3-6}          &       & PSNR  & SSIM  & PSNR  & SSIM \\
    \midrule
    \multirow{3}[3]{*}{1400-1200M-100H-100L} & MFDNet & 31.93$\pm$0.05 & 0.923$\pm$0.002 & 34.12$\pm$0.02 & 0.949$\pm$0.001 \\
\cmidrule{2-6}          & Restormer & 32.48$\pm$0.06 & 0.927$\pm$0.002 & 34.04$\pm$0.01 & 0.948$\pm$0.000 \\
\cmidrule{2-6}          & MPRNet & 31.69$\pm$0.07 & 0.916$\pm$0.001 & 33.54$\pm$0.02 & 0.946$\pm$0.001 \\
    \bottomrule
    \end{tabular}%
    }
  \label{tab:stability}%
  \vspace{-0.3cm}
\end{table}

\textbf{Validation on hyper-parameter $\lambda$.}
We conduct ablation studies to verify the hyper-parameter $\lambda$ of balancing the two loss terms: the interleave loss, and consistency loss. The experiment results are illustrated in Table~\ref{tab:hyper}. We found that excessively large or small hyper-parameter can lead to an imbalance between the two losses during optimization, resulting in a decrease in memory and generalization performance. Considering the trade-off, we ultimately set $\lambda$ to 1 through a comprehensive search, thereby achieving optimal memory and generalization performance.

{
\textbf{Validation on framework stability.}
To validate the stability of the proposed framework, we conduct experiments using all three de-raining networks and perform validation under one four-dataset stream (1400-1200M-100H-100L). We perform 10 independent runs and report the mean and standard deviation of PSNR and SSIM for both memory and generalization performance. As shown in Tab.~\ref{tab:stability}, the standard deviation of PSNR remains below 0.10 dB, and that of SSIM is less than 0.003 across all settings. These variations are considerably smaller than the observed performance differences between CLGID and competing methods, indicating that the influence of randomness is negligible. The results confirm that our framework is stable.}

{
\textbf{Visualization of loss convergence.}
To verify the convergence behavior of the proposed framework, we provide the training loss curve under a representative setting (MFDNet + 1400-1200M-100H-100L), as shown in Fig.~\ref{Loss}. The curve clearly demonstrates that the training process exhibits smooth and stable convergence. Within each dataset training phase, the loss decreases steadily, showing consistent optimization. Moreover, at the transition between datasets, there is no sign of abrupt increase, conflict, or collapse in the loss, indicating that the framework handles dataset shifts in a stable manner.}

\section{Conclusion}
\label{section:Conclusion}
This paper presents a new generalized de-raining framework, CLGID, which empowers de-raining networks to accumulate knowledge from increasingly abundant de-raining datasets, rather than relying solely on a static dataset, thereby constantly improving their ability to generalize to unseen real-world scenes. Our inspiration stems from the human brain's complementary learning system, which enables humans to constantly learn and memorize a stream of perceived events and gradually acquire the generalization ability to unseen situations across memorized events. This remarkable human ability closely aligns with our research goals. Therefore, we endeavor to borrow the mechanisms of the complementary learning system into our framework to achieve our goals. Extensive experiments are conducted to validate our approach's effectiveness. Our CLGID empowers de-raining networks to effectively accumulate de-raining knowledge from a stream of datasets and constantly enhance their generalization performance in unseen real-world rainy scenes.

\begin{figure*}[htbp]
\begin{center}
\includegraphics[width=0.90\linewidth]{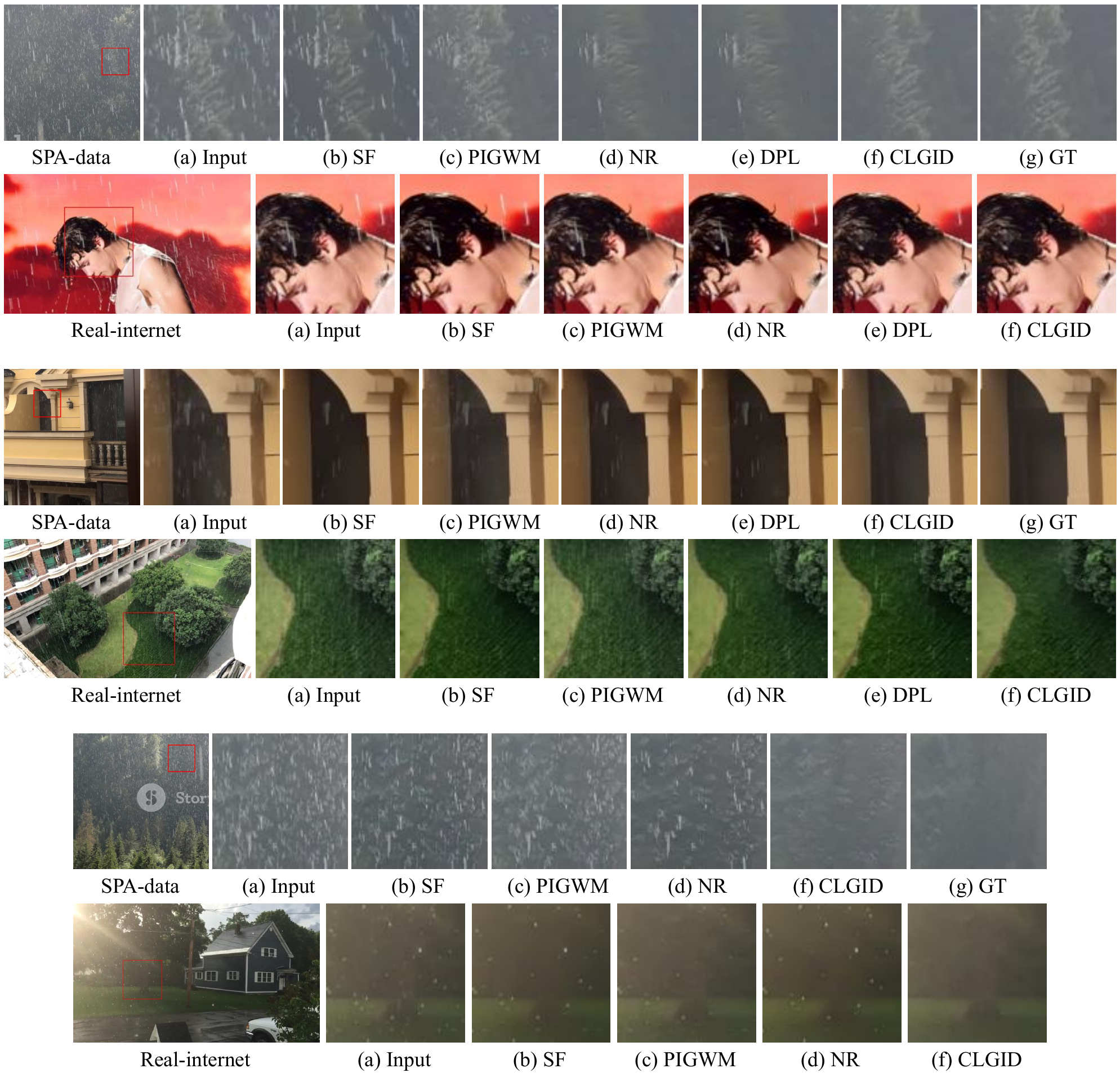}
\end{center}
\vspace{-0.3cm}
\caption{{Visual quality comparisons of different methods on SPA-data~\cite{wang2019spatial} and Real-internet~\cite{wang2019spatial}. From top to bottom: MFDNet~\cite{wang2023multi}, Restormer~\cite{zamir2022restormer}, and MPRNet~\cite{zamir2021multi} are used as the de-raining networks. Ground truth is available for SPA-data but not for Real-internet. Note that DPL is not applicable to non-transformer-based networks such as MPRNet.}
}
\vspace{-0.4cm}
\label{fig8}
\end{figure*}

\appendices
{
\section{Framework Complexity Analysis}
\label{complexity}
We analyze how the cost in terms of parameters, computation, and time scales with the number of integrated datasets $N$.
Let $D_n$ denote the $n$-th dataset, with $M_n$ image pairs. The CLGID training framework consists of three core stages at each step: training a GAN $G_n$ on $D_n$, generating replay data $\hat{D}_n$ from earlier GANs $\{G_1,\cdots,G_{n-1}\}$, and training the de-raining network on $D_n \cup \hat{D}_n$.}

{\noindent{\textbf{GAN Complexity Analysis}}}

{\noindent\textbf{Parameter cost}: Each new dataset adds one GAN. If a single GAN contains $P_G$ parameters, total parameter cost scales linearly:  
\[
P_{\text{GAN}} = N \cdot P_G
\]
\noindent\textbf{Training FLOPs and time}: Assume GANs are trained for $E_G$ epochs with per-batch ($B_G$) FLOPs cost $F_G^{\text{Train}}$ and time cost $t_G^{\text{Train}}$. Then for each stage:
\[
\text{FLOPs}_\text{GAN}^{(n)} = E_G \cdot \frac{M_n}{B_G} \cdot F_G^{\text{Train}},
\]
\[
T_\text{GAN}^{(n)} = E_G \cdot \frac{M_n}{B_G} \cdot t_G^{\text{Train}}.
\]
Accumulated over $N$ datasets:
\[
\text{FLOPs}_\text{GAN} = \sum_{n=1}^N E_G \cdot \frac{M_n}{B_G} \cdot F_G^{\text{Train}},
\]
\[
T_\text{GAN} = \sum_{n=1}^N E_G \cdot \frac{M_n}{B_G} \cdot t_G^{\text{Train}}.
\]}

{\noindent{\textbf{Replay FLOPs and time}}: At stage $n$, replay dataset $\hat{D}_n$ is generated to match the size of the current dataset $T_n$, i.e., $|\hat{D}_n| = M_n$. The samples are drawn by uniformly sampling from each of the $n - 1$ GANs. Each sample requires FLOPs $F_R$ and time $T_R$:
\[
\text{FLOPs}_\text{Replay}^{(n)} = M_n \cdot F_R, \quad T_\text{Replay}^{(n)} = M_n \cdot T_R.
\]
Accumulated over $N$ datasets:
\[
\text{FLOPs}_\text{Replay} = \sum_{n=1}^N M_n \cdot F_R, \quad T_\text{Replay} \sum_{n=1}^N M_n \cdot T_R
\]}

{\noindent{\textbf{De-raining Network Complexity Analysis}}}

{\noindent\textbf{Parameter cost}: The de-raining backbone remains fixed throughout training. Let it have $P_D$ parameters:
\begin{equation}
P_{\text{D-net}} = P_D \; \text{(constant w.r.t. } N\text{)}.
\end{equation}}

{\noindent\textbf{Training FLOPs and time}:
Let $F_D^\text{Train}$ and $t_D^\text{Train}$ be the FLOPs and time cost for per-batch ($B_D$) forward–backward pass. With $E_D$ training epochs:
\[
\text{FLOPs}_\text{D-net}^{(n)} = E_D \cdot \frac{M_n}{B_D} \cdot F_D^{\text{Train}},
\]
\[
T_\text{D-net}^{(n)} = E_D \cdot \frac{M_n}{B_D} \cdot t_D^{\text{Train}}.
\]
Accumulated over $N$ datasets:
\[
\text{FLOPs}_\text{D-net} = \sum_{n=1}^N E_D \cdot \frac{M_n}{B_D} \cdot F_D^{\text{Train}},
\]
\[
\text{T}_\text{D-net} = \sum_{n=1}^N E_D \cdot \frac{M_n}{B_D} \cdot t_D^{\text{Train}}.
\]}

{\section{The proof of the total replay cost in GAN-replayed data reuse}
\label{math}
Suppose the dataset sizes are upper bounded: $M_n \leq M_{\max}$, and denote $M_{\min} = \min_n M_n > 0$. Then we can derive an upper bound:
\[
\Delta_{i,n} \leq \max\left(0, \frac{M_{\max}}{n-1} - \frac{M_{\min}}{n-2} \right).
\]
More importantly, the dominant term across all $n$ is:
\[
\Delta_{n-1,n} = \frac{M_n}{n-1} \leq \frac{M_{\max}}{n-1}.
\]
Therefore, the total cost satisfies:
\[
C_N \leq \sum_{n=2}^{N} \left((n-2) \cdot \varepsilon_n + \frac{M_{\max}}{n-1} \right),
\]
where 
\[
\varepsilon_n = \max\left(0, \frac{M_{\max}}{n-1} - \frac{M_{\min}}{n-2} \right).
\]
Note that $\varepsilon_n$ vanishes for sufficiently large $n$.
Specifically, solving
\[
\frac{M_{\max}}{n-1} \leq \frac{M_{\min}}{n-2} \Longleftrightarrow \frac{M_{\max}}{M_{\min}} \leq \frac{n-1}{n-2},
\]
gives a threshold 
\[
N_0 = \left\lceil 1 + \frac{M_{\max}}{M_{\min} - M_{\max}} \right\rceil.
\]
above which $\varepsilon_n = 0$ for all $n \geq N_0$. Thus,
\[
\sum_{n=2}^{N} (n-2) \varepsilon_n \leq \sum_{n=2}^{N_0} (n-2) \varepsilon_n \triangleq C_0,
\]
where $C_0$ is a finite constant independent of $N$. Therefore, we can treat the first summation as a constant. The second term forms a harmonic sum:
\[
\sum_{n=2}^{N} \frac{M_{\max}}{n-1} = M_{\max} \sum_{k=1}^{N-1} \frac{1}{k} = M_{\max} \cdot H_{N-1},
\]
where $H_{N-1} \leq \ln(N-1) + 1$. Therefore,
\[
C_N = \mathcal{O}(M_{\max} \log N).
\]}


%





\ifCLASSOPTIONcaptionsoff
  \newpage
\fi



%

\bibliographystyle{IEEEtran}
\bibliography{bibtex/bib/IEEEabrv}



%








\end{document}